\documentclass[letterpaper]{article} 
\usepackage{aaai24}  
\usepackage{times}  
\usepackage{helvet}  
\usepackage{courier}  
\usepackage[hyphens]{url}  
\usepackage{graphicx} 
\usepackage{natbib}  
\usepackage{caption} 
\usepackage{algorithm}
\usepackage{algorithmic}

\usepackage{newfloat}
\usepackage{listings}
\DeclareCaptionStyle{ruled}{labelfont=normalfont,labelsep=colon,strut=off} 
\floatstyle{ruled}
\newfloat{listing}{tb}{lst}{}
\floatname{listing}{Liåsting}
\usepackage[utf8]{inputenc} 
\usepackage{booktabs}       
\usepackage{amsfonts}       
\usepackage{nicefrac}       
\usepackage{microtype}      
\usepackage{xcolor}         
\usepackage{times}
\usepackage{latexsym}
\usepackage{amsmath,amsfonts,bm}
\usepackage{graphics}
\usepackage{subcaption}
\usepackage{multirow}
\usepackage[para]{threeparttable}
\usepackage{tabularx}
\usepackage{xspace}
\usepackage{xcolor,colortbl}
\usepackage{pifont}
\usepackage{spverbatim}
\usepackage{adjustbox}
\usepackage{capt-of}
\usepackage{ragged2e}
\usepackage{multicol}
\usepackage{makecell}
\definecolor{darkgreen}{RGB}{0, 170, 0} 
\newcommand{\cmark}{\ding{51}}
\newcommand{\xmark}{\textcolor{red}{\ding{55}}}
\newcommand{\xgmark}{\textcolor{darkgreen}{\ding{55}}}
\newcommand{\ICIL}{\textsc{TAPP}\xspace}
\DeclareMathOperator*{\argmax}{arg\,max}
\usepackage{tcolorbox}

\title{Investigating the Effectiveness of Task-Agnostic Prefix Prompt \\for Instruction Following}
\author{%
  Seonghyeon Ye{\textsuperscript{1}}\thanks{~~Work done while interning at LG AI Research.} \quad Hyeonbin Hwang{\textsuperscript{1}}\quad Sohee Yang{\textsuperscript{1,2}} \\ {\bf \quad Hyeongu Yun{\textsuperscript{3}} \quad Yireun Kim{\textsuperscript{3}} \quad Minjoon Seo{\textsuperscript{1}}} \\
  {\textsuperscript{1}KAIST \quad \textsuperscript{2} UCL \quad \textsuperscript{3} LG AI Research} \\
  \texttt{seonghyeon.ye@kaist.ac.kr}
  }

\usepackage{bibentry}

\begin{document}
\maketitle

\begin{abstract}

In this paper, we present our finding that prepending a Task-Agnostic Prefix Prompt (TAPP) to the input improves the instruction-following ability of various Large Language Models (LLMs) during inference. \ICIL is different from canonical prompts for LLMs in that it is a \textit{fixed} prompt prepended to the beginning of every input regardless of the target task for zero-shot generalization. We observe that both base LLMs (i.e. not fine-tuned to follow instructions) and instruction-tuned models benefit from \ICIL, resulting in 34.58\% and 12.26\% improvement on average, respectively. This implies that the instruction-following ability of LLMs can be improved during inference time with a fixed prompt constructed with simple heuristics. 
We hypothesize that \ICIL assists language models to better estimate the output distribution by focusing more on the instruction of the target task during inference. In other words, such ability does not seem to be sufficiently activated in not only base LLMs but also many instruction-fine-tuned LLMs\footnote{All experiments are reproducible from \url{github.com/seonghyeonye/TAPP}.}.
\end{abstract}

\section{Introduction}
Large Language Models (LLMs) have demonstrated the ability to follow user instructions through approaches such as instruction tuning or reinforcement learning from human feedback (RLHF) \citep{sanh2021multitask, wei2021finetuned, wang2022benchmarking, ouyang2022training, min-etal-2022-metaicl, chung2022scaling, ye2022guess, bai2022training, askell2021general}. However, previous work mainly has focused on fine-tuning-based approaches to enhance the instruction-following ability of LLMs where the model is fine-tuned on various tasks with instructions, requiring multiple backpropagation processes and necessitating access to the model weights which limits its applicability to proprietary models. 

In this paper, we present and analyze our finding that prepending a \textbf{T}ask-\textbf{A}gnostic \textbf{P}refix \textbf{P}rompt (TAPP) that is determined by simple heuristics during inference significantly enhances the instruction-following ability of LLMs across various tasks for both open-sourced and proprietary models \citep{zhang2022opt, brown2020language, gpt-j, black2022gpt}. Specifically, \ICIL consists of multiple cross-task demonstrations where each demonstration is a concatenation of an instruction, input, and output instance of a task. Note that \ICIL is different from canonical task-specific prompts in that it is a fixed prefix prompt that is prepended regardless of the target task for zero-shot generalization.

We first observe that \ICIL significantly enhances the instruction-following performance of various base LLMs that are not fine-tuned to follow instructions. Notably, even smaller LLMs with \ICIL outperform much larger language models without \ICIL, such as the 6B-sized GPT-J with \ICIL outperforming 30 times larger 175B-sized GPT-3 Davinci without \ICIL. Second, we show that applying \ICIL on top of instruction-fine-tuned LLMs also improves the performance, boosting the performance of one of the strongest instruction-following LLMs (text-davinci-003) by 9.3\%. This indicates that the effect of \ICIL during inference is complementary to the effect of instruction fine-tuning. Moreover, we demonstrate that prepending \ICIL to target task demonstrations also improves performance, implying that \ICIL also enhances few-shot in-context learning during inference.

\begin{figure*}
    \centering
    \includegraphics[width=0.8\linewidth]{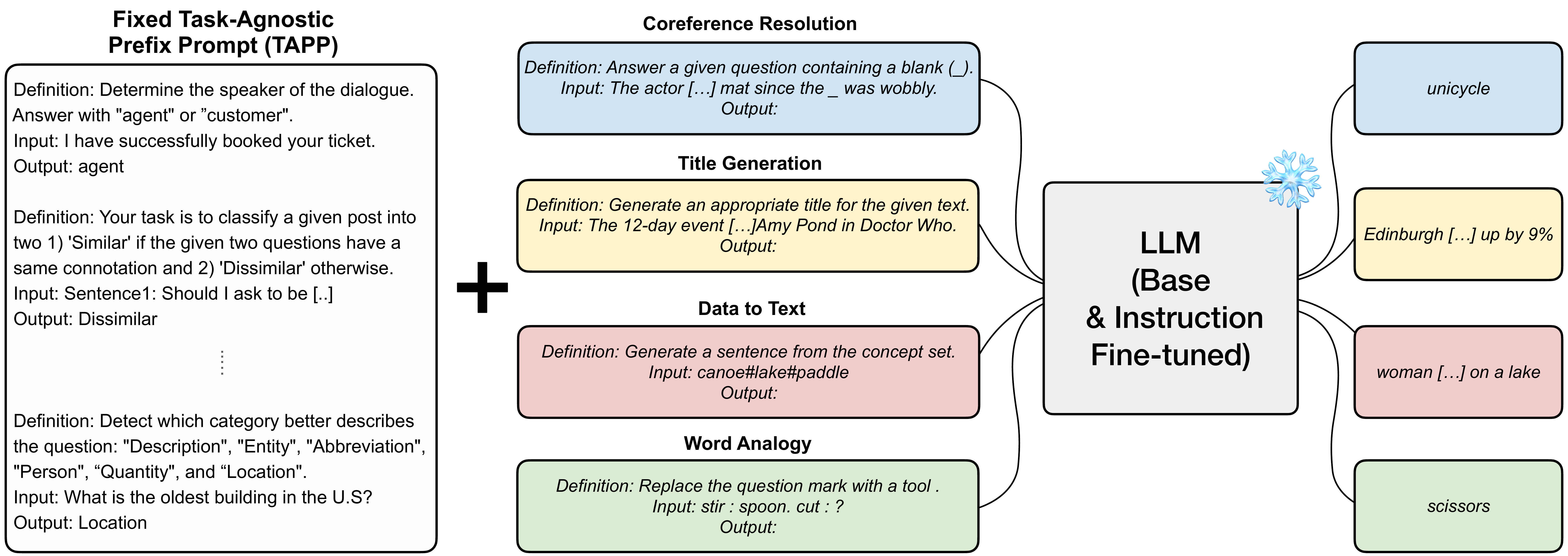}
    \caption{Overview of Task-Agnostic Prefix Prompt (\ICIL). We construct a fixed set of demonstrations consisting of instruction, input, and output instances to evaluate base and instruction-fine-tuned LLMs for all tasks. The task categories included in the demonstrations are strictly held-out and from the tasks being evaluated, ensuring a zero-shot setting.}
    \label{fig:figure_1}
    \vspace{-2mm}
\end{figure*}

Our analysis shows that \ICIL performs best when the prefix prompt consists of demonstrations of classification tasks that include explicit answer choice in the instruction (e.g., expression of \textit{``agent" or ``customer"} in Figure \ref{fig:figure_1}). This holds true even when the target task is a generation task, which contrasts with the findings of the previous studies that it is crucial to retrieve a set of prompts that are similar to the target task \citep{rubin2021learning, liu2021makes}. We also observe that the performance does not degrade significantly even if the input distribution of each demonstration of \ICIL is corrupted. Based on these two observations, we hypothesize that during inference of \ICIL, LLMs learn the correspondence between the answer choice included in instruction and the output of each demonstration of \ICIL. Through this hypothesis, we suggest that the role of \ICIL is to help LLMs \textit{focus} on the target instruction to better estimate the output distribution of the target task. This also implies that this ability does not seem to be sufficiently activated in both base LLMs and instruction-tuned LLMs, leaving further investigation as future work.

\section{Related Works}\label{sec:rel-works}

\paragraph{Inference-time Task Adaptation} 
In-context learning is one of the most widely known gradient-free task adaptation approaches during inference. Language models pretrained to predict the next token autoregressively possess the ability to adapt to the target tasks when conditioned on only a few task-specific training examples without gradient update \citep{brown2020language, chowdhery2022palm, akyurek2022learning, von2022transformers, garg2022can,dai2022can}. 
However, few-shot in-context learning requires access to target task demonstrations for each task, implying that the user has to take the effort of generating the demonstrations for each task by themselves. 
To address this issue, \citet{lyu2022z} propose Zero-shot In-Context Learning method (Z-ICL), retrieving relevant sentences from an external corpus and assigning random labels to construct demonstrations for classification target tasks. However, Z-ICL is only applicable for single-sentence classification tasks and tasks that only have single-word answer choices. Also, Z-ICL assumes that the output distribution of the task is given. In contrast, our work observes the effect of task-agnostic prefix prompts without any restrictions on the type of the downstream task or the necessity of additional information about the task, which makes the approach applicable even for real-time scenarios.
\paragraph{Instruction-Following LLMs}
Recent works have shown that fine-tuning-based instruction learning, e.g., instruction tuning or RLHF, can boost the capability of LLMs to follow instructions or align to human preferences \citep{sanh2021multitask, wei2021finetuned, wang2022benchmarking, chung2022scaling, min-etal-2022-metaicl, ye2022guess, ouyang2022training, bai2022training, chatgpt}. These works have demonstrated that the effect of instruction fine-tuning can be maximized by scaling the size of the base model or by training on a more diverse set of tasks. However, whether the instruction following ability of LLMs is newly obtained through instruction tuning or is already obtained during pretraining is under-explored. \citet{wang2022self, honovich2022unnatural} show that downstream tasks generated by LLMs themselves which contain noisy instances can actually be good training instances for instruction tuning, implying that LLMs are already somewhat aware of instructions. We extend this hypothesis that base LLMs already have the capability to follow instructions by showing that applying \ICIL regardless of the target task without performing any backpropagation, i.e., using the base model checkpoint without any gradient update, improves the performance on the target downstream tasks.

\section{Task-Agnostic Prefix Prompt}  

\ICIL consists of cross-task demonstrations where each is a concatenation of instruction, input, and output instance, as shown in Figure \ref{fig:figure_1}. The exact prompt is provided in Appendix \ref{appen:prom_i2c}.
In this section, we explain the rules we have used to construct \ICIL. Also, we mention the advantages of applying \ICIL during the inference of LLMs for zero-shot task generalization.

\subsection{\ICIL Construction}
\begin{figure*}
    \centering
    \includegraphics[width=0.8\linewidth]{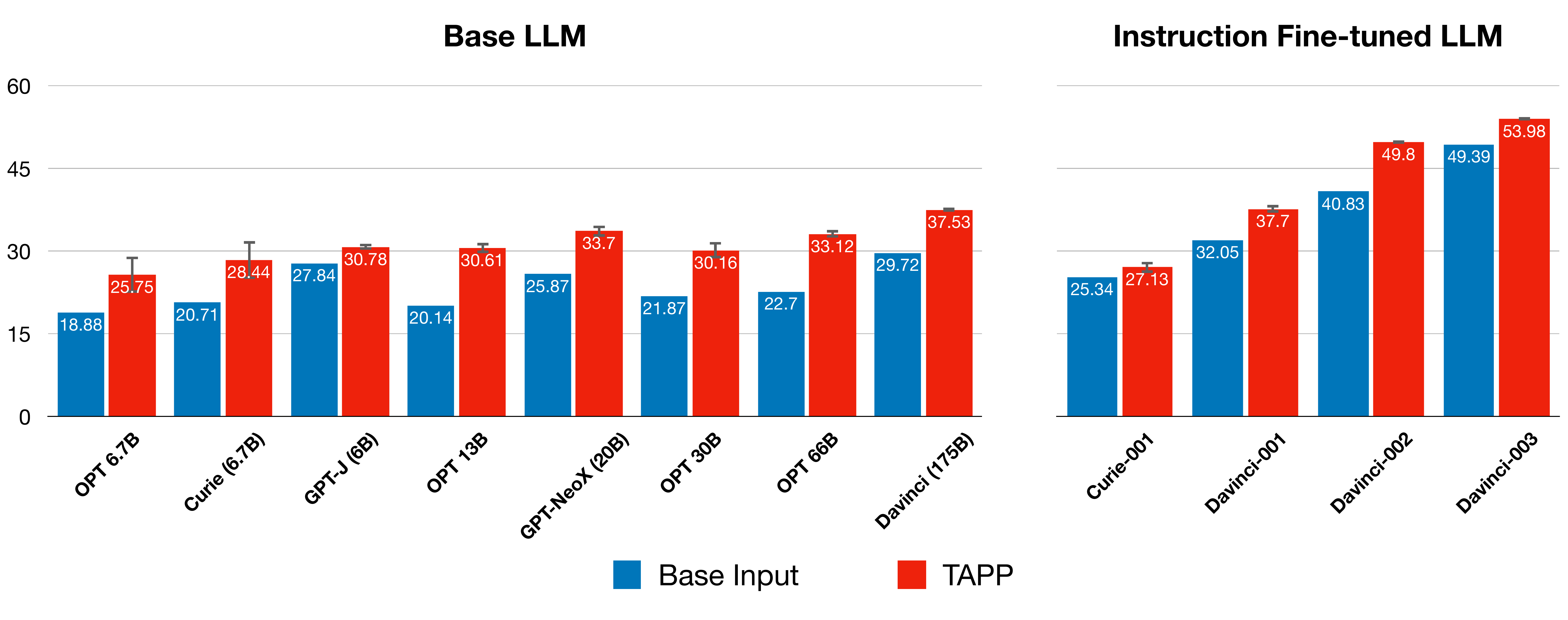}
    \caption{Average performance of 119 evaluation tasks on \textsc{SuperNI} benchmark. \ICIL is effective for both base and instruction-fine-tuned LLMs. We report the mean score of three random seeds for different demonstration sets for \ICIL and the error bars of standard deviation. We also perform an evaluation using INSTRUCTSCORE \citep{xu2023instructscore} in Appendix \ref{appen:instructscore} and provide the full demonstration sets in Appendix \ref{appen:prom_i2c}}
    \label{fig:main_result}
\vspace{-2mm}
\end{figure*}

We select $K$ tasks as demonstrations for \ICIL from a task pool containing a total of $N$ tasks, with each task instance consisting of an instruction, input, and output\footnote{Unless specified, we set $K=8$ as default.}. We apply some simple heuristics to first filter the task set, randomly sample a single instance per filtered task set, and lastly, sample $K$ instances all corresponding to different tasks. 
The rules are as follows:
\begin{enumerate}
    \item Task Types: We only sample from classification tasks that explicitly include an answer choice in the instruction (e.g., \textit{``agent" or ``customer"} in Figure \ref{fig:figure_1}). We hypothesize that including the answer choice in the instruction might assist LLMs to follow instructions during inference.
    \item Answer Choice Overlap: We ensure that the answer choices do not overlap between demonstrations. We expect that the overlap of answer choices leads to LLMs copying the labels of the demonstrations, similar to the copying effect during inference of LLMs \citep{lyu2022z}.
    \item Maximum Length: We restrict the length of the concatenation of instruction, input, and output instance for each demonstration to 256 tokens by a maximum considering the maximum sequence length\footnote{Because we mainly experiment on 175B-sized GPT-3, we set the default maximum input sequence as 2048.}. 
    \item Ordering: We order the demonstrations by the number of answer choices for each task in ascending order. For demonstrations having the same number of answer choices, we sort by demonstration length in ascending order. 

\end{enumerate}
We provide a detailed analysis and ablation of these heuristics in Section \ref{sec:ablation}, justifying our design of rules.
\subsection{\ICIL for Zero-Shot Task Generalization}

After randomly sampling $K$ tasks from a set of tasks that satisfy the criteria and ordering them by the criterion, we construct a fixed set of demonstrations $M = [M_1, M_2, ..., M_K]$ (\ICIL) and prepend it on the concatenation of instruction ($I_t$) and $i$-th input instance ($x_{ti}$) of the target task $t$.
The response ($y_{ti}$) of the model parameterized by $\theta$ is calculated as follows:

\begin{equation}
    \argmax P(y_{ti}|M, I_t,x_{ti} ; \theta)
\end{equation}
where $M$ is invariant regardless of the target task $t$ and $K$ is the number of demonstrations. We ensure that the $K$ tasks comprising the demonstration set of \ICIL are strictly held-out from the target task $T$ in order to measure the effect of \ICIL for zero-shot task generalization.

It is worth noting that \ICIL is different from canonical task-specific prompts which usually vary depending on the target task. \ICIL is a fixed prefix prompt that can be prepended to any target task without any restriction, being easily reproducible and widely applicable as described in Section~\ref{sec:rel-works}. 

Also, \ICIL does not require any additional information during inference such as the task category information or the output distribution of the target task, unlike task-specific prompts such as few-shot prompting or the approach of \citet{lyu2022z}.

\section{Experiments}
\subsection{Experimental Setup}

We construct the demonstrations for \ICIL by utilizing English training tasks of \textsc{Super-NaturalInstructions (SuperNI)} benchmark \citep{wang2022benchmarking} as the task pool, which includes 756 tasks in total. To evaluate the effectiveness of \ICIL, we use the held-out tasks from \textsc{SuperNI} for testing, which consists of 119 tasks across 12 different categories, including free-form generation, word relation reasoning, and various classification tasks. We select \textsc{SuperNI} as our evaluation benchmark because it offers a diverse set of tasks with varying levels of complexity. Each task has 100 instances, and we exclude instances that exceed the maximum sequence length, resulting in a total of 11,802 instances. We use different evaluation metrics for each task, such as Exact Match for classification or single-word prediction tasks and ROUGE-L for free-form generation tasks, following the metric used in \citet{wang2022benchmarking}. 
We provide the list of 12 evaluation task categories in Appendix \ref{appen:full_result} and more detailed evaluation settings in Appendix \ref{appen:evaluation_setting}.

\paragraph{Model Types}
We evaluate 4 LLMs with various model sizes: 1) GPT-3 \citep{brown2020language}, 2) OPT \citep{zhang2022opt}, 3) GPT-NeoX \citep{black2022gpt}, and 4) GPT-J \citep{gpt-j}\footnote{From preliminary experiments, we observe that applying \ICIL harms the performance for OPT-IML \citep{iyer2022opt} and FLAN-T5 \citep{chung2022scaling} due to the characteristics of each model. We provide more discussion in Appendix \ref{appen:opt_iml}.}. 
For GPT-3, we evaluate not only the base LLM but also evaluate LLMs that are fine-tuned to follow instructions and aligned to human preferences through reinforcement learning \citep{ouyang2022training}. We evaluate the performance of GPT-3 models with sizes of 6.7B and 175B. For OPT, we evaluate models with 6.7B, 13B, and 30B parameters, while for GPT-NeoX and GPT-J, we evaluate models with 20B and 6B parameters, respectively\footnote{We do not evaluate on GPT-3.5 \citep{chatgpt} or GPT-4 \citep{openai2023gpt4} since the model details such as the model size or architecture are not known.}.
\begin{figure*}
    \centering
    \begin{subfigure}[b]{0.8\textwidth}
        \includegraphics[width=\textwidth]{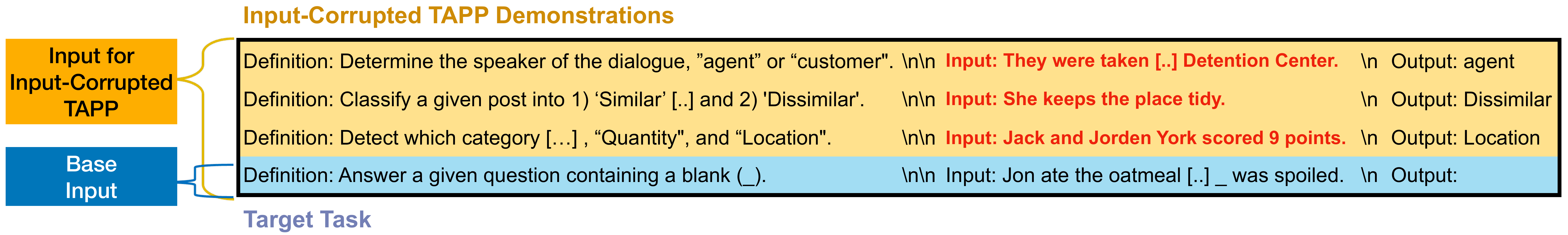}
    \end{subfigure}
    \begin{subfigure}[b]{0.8\textwidth}
        \includegraphics[width=\textwidth]{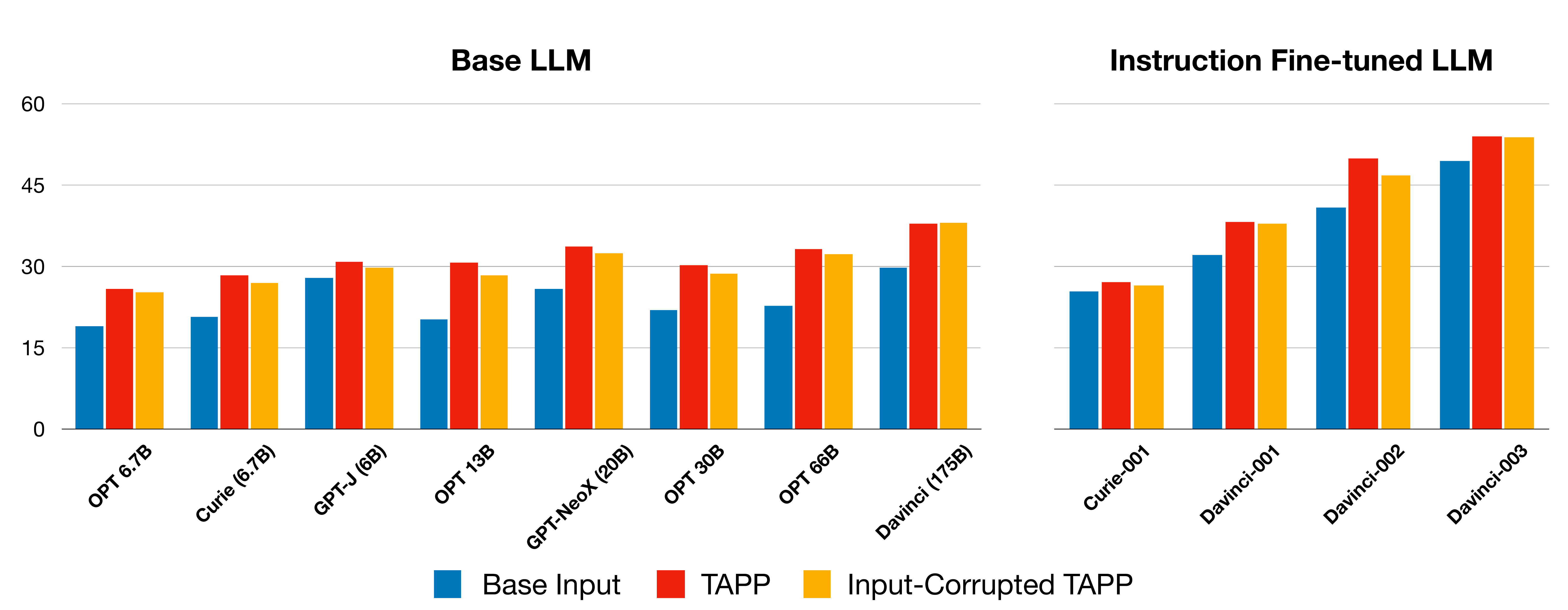}
    \end{subfigure}
    \caption{(Top) Example of Input-corrupted \ICIL, where we corrupt the input instance distribution of the demonstrations. (Bottom) Comparison with Base Input, \ICIL, and Input-corrupted \ICIL. For most of the models, input distribution corruption does not harm the performance much. We report the mean score of three random seeds for different demonstration sets for \ICIL. We report a result of a single seed for 175B-sized models due to inference costs. We provide the full demonstration sets in Appendix \ref{appen:i3c}.}
    \label{fig:i3c}
\vspace{-2mm}
\end{figure*} 
\subsection{Results}

\paragraph{Various base LLMs benefit from \ICIL.}
As shown in the left part of Figure \ref{fig:main_result}, Task-Agnostic Prefix Prompt (\ICIL) consistently improves the performance of base LLMs (i.e not fine-tuned with instructions) across all model scales, resulting in over 50\% performance increase for OPT-13B. Using this fixed prompt, smaller models can outperform much larger models without \ICIL (Base Input). Specifically, the 6B-sized GPT-J model with \ICIL exceeds 30 times larger 175B-sized GPT-3 model without \ICIL. This shows that \ICIL can improve the ability of base LLMs to follow instructions without fine-tuning or backpropagation. Moreover, we observe the gain from \ICIL during inference can outperform the gain from instruction tuning by comparing the performance of \ICIL applied to GPT-3 model without instruction tuning (davinci) and the base input setting of the instruction-tuned GPT-3 model (text-davinci-001).

\paragraph{The gain from \ICIL is complementary to instruction fine-tuning.}
As shown in the right part of Figure \ref{fig:main_result}, we observe that \ICIL improves the performance of LLMs fine-tuned through instruction tuning or RLHF, especially for models over 100B parameters. This implies that instruction fine-tuning alone might be sometimes insufficient for larger models and pretending a fixed prefix prompt can improve the instruction following ability orthogonally. In particular, we observe a significant performance improvement for text-davinci-002 (175B), outperforming an RLHF-tuned model text-davinci-003 with base input. Also, we demonstrate that the most powerful model (text-davinci-003) also benefits from \ICIL by 9.3\%, achieving the best performance. We leave detailed analysis on more diverse instruction-fine-tuned models as future work. 

\paragraph{Input Corruption of \ICIL does not harm the performance much.}

In Figure \ref{fig:i3c}, we observe that corrupting the distribution of input instances for each demonstration for \ICIL does not harm the performance much, similar to the observation in \citet{min2022rethinking} for few-shot in-context learning.
Instead of perturbing the input-output correspondence, done in \citet{min2022rethinking}, we perturb the input distribution \textit{itself}, which is a setting where there are more corruptions as shown at the top of Figure \ref{fig:i3c}. Following \citet{min2022rethinking}, we use CC-News \citep{Hamborg2017} as an external corpus to replace the ground truth input instance with random sentences that have a similar length to the original input instance. As shown in the bottom of Figure \ref{fig:i3c}, corrupting the input instance distribution of each demonstration does not harm the performance much across most model scales. This is in line with the observations made in previous works that LLMs do not make full use of all the information provided to them \citep{min2022rethinking, webson2021prompt, madaan2022text, wang2022towards}. Interestingly, unlike few-shot in-context learning where corrupting the input distribution itself leads to significant performance degradation, we demonstrate that not only the input-output correspondence does not matter, but also the input instance distribution matters little for \ICIL. 
\label{sec:results}

\section{Analysis}
\label{sec:ablation}
In this section, we provide additional experiments and investigate the factors that make \ICIL effective. We evaluate on base GPT-3 175B checkpoint (davinci) and evaluate on a single task per task category, resulting in a total of 12 tasks due to inference cost issues\footnote{We select a single task per task category with a significant discrepancy between the lower bound and upper bound performance across davinci, text-davinci-001, 002, 003 models to see the tendency more clearly.}.

\subsection{Additional Experiments}

\begin{table}[]
\centering
\fontsize{8.5}{11}\selectfont

\begin{tabular}{l|ccccc}
 & Category   & Output &  Task & AVG\\ \midrule
Base (No PP) & \xgmark  & \xgmark & \xgmark & 29.66\\
TAPP & \xgmark & \xgmark & \xgmark & \textbf{44.24}\\
Nearest PP & \xgmark & \xgmark & \xgmark & 44.16\\
Category PP & \cmark & \xgmark & \xgmark & 42.43\\
Output PP & \xgmark & \cmark & \xgmark & 34.34\\ \midrule \midrule
Few-shot ICL & \cmark & \cmark & \cmark & 56.65\\
+ \ICIL & \cmark & \cmark & \cmark & \textbf{60.21}\\

\end{tabular}

\caption{We compare the performance of \ICIL with different strategies to construct task-specific Prefix Prompts (PP): Nearest PP, Category PP, and Output PP. Unlike other approaches, \ICIL is fixed regardless of the target task and does not require any information about the task category, output, or target task. Additionally, we observe prepending \ICIL to target task demonstrations (Few-shot ICL) enhances the performance.}
\label{table:baseline}
\vspace{-5mm}
\end{table}

\paragraph{Comparison with Task-specific Prefix Prompts}
We compare the performance of \ICIL with other prefix prompts that are task-specific, meaning that the prefix prompt depends on the target task rather than being fixed. We compare with three approaches: Nearest PP, Category PP, and Output PP. First, for Nearest PP, we construct the prefix prompt by retrieving top-$K$ similar instances for each target task from training tasks of \textsc{SuperNI} using SimCSE \citep{gao2021simcse} search tool, similar to the setting of \citet{lyu2022z}\footnote{Note that the original setting of \citet{lyu2022z} is only applicable for single sentence classification tasks and for tasks that have single word answer choices. Therefore, the method cannot be directly compared to benchmarks that include a diverse collection of tasks such as \textsc{SuperNI}.}. Second, for Category PP, we construct the prefix prompt by randomly sampling demonstrations from the task that belongs to the same task category (e.g., question answering), from the evaluation tasks of \textsc{SuperNI} benchmark but excluding the same task, assuming that the task category information is provided during inference. Third, for Output PP, we utilize the output label of few-shot demonstrations of the target task while corrupting the input distribution of each demonstration, following the input corruption setting of \citet{min2022rethinking}. This setting is equivalent to providing only the output distribution through demonstrations.

Results in Table \ref{table:baseline} show that \ICIL is comparable to or outperforms other task-specific prefix prompts. First, we find that Nearest PP does not outperform \ICIL. This indicates that using an external search tool to find similar demonstrations for each target task might not help much. Second, Category PP slightly underperforms \ICIL because constructing cross-task demonstrations that are similar to the target task sometimes leads to copying the output of the demonstration, similar to the copying effect observed in \citet{lyu2022z}. Lastly, we observe that Output PP significantly underperforms \ICIL. Although Output PP outperforms \ICIL for classification tasks (36.83 vs 43.67), it significantly underperforms for generation tasks (51.93 vs 24.98). We hypothesize that this is because while the correspondence between input and output for each demonstration is less crucial for classification tasks \citep{min2022rethinking}, the correspondence is important for generation tasks, giving a distracting signal to the LLM if the correspondence is not matched \citep{shi2023large}. 
Through these results, we observe that \ICIL shows comparable or better performance compared to task-specific prefix prompts while not requiring any additional information or search tools.
\paragraph{Orthogonality with Few-shot In-Context Learning}

From the result of Figure \ref{fig:main_result}, we have observed that \ICIL enhances the performance of instruction-fine-tuned LLMs. Here, we investigate if \ICIL also enhances few-shot in-context learning, which assumes that in addition to category and output information, the target task information is provided through demonstrations of the target task. We use 4-shot few-shot demonstrations for Few-shot ICL and prepend 4 demonstrations for \ICIL to fit the input into the maximum sequence length. As shown in Table \ref{table:baseline}, prepending \ICIL to Few-shot ICL boosts the performance, implying that \ICIL can also enhance few-shot in-context learning through a fixed prefix prompt. Additionally, we observe that prepending 4-shot \ICIL to 4-shot Few-shot ICL setting largely reduces the performance gap between 8-shot Few-shot ICL without \ICIL (60.21 vs 61.71). 
This suggests the advantage of \ICIL in real-time LLM serving scenarios: while the users can save the effort of manually constructing twice more task-specific demonstrations for each task, they can achieve similar performance by simply prepending \ICIL to the input.

\paragraph{Comparison with Machine-Generated Prefix Prompts}
\label{sec:additional}
\begin{figure}[ht!]
\centering
\includegraphics[width=0.3\textwidth]{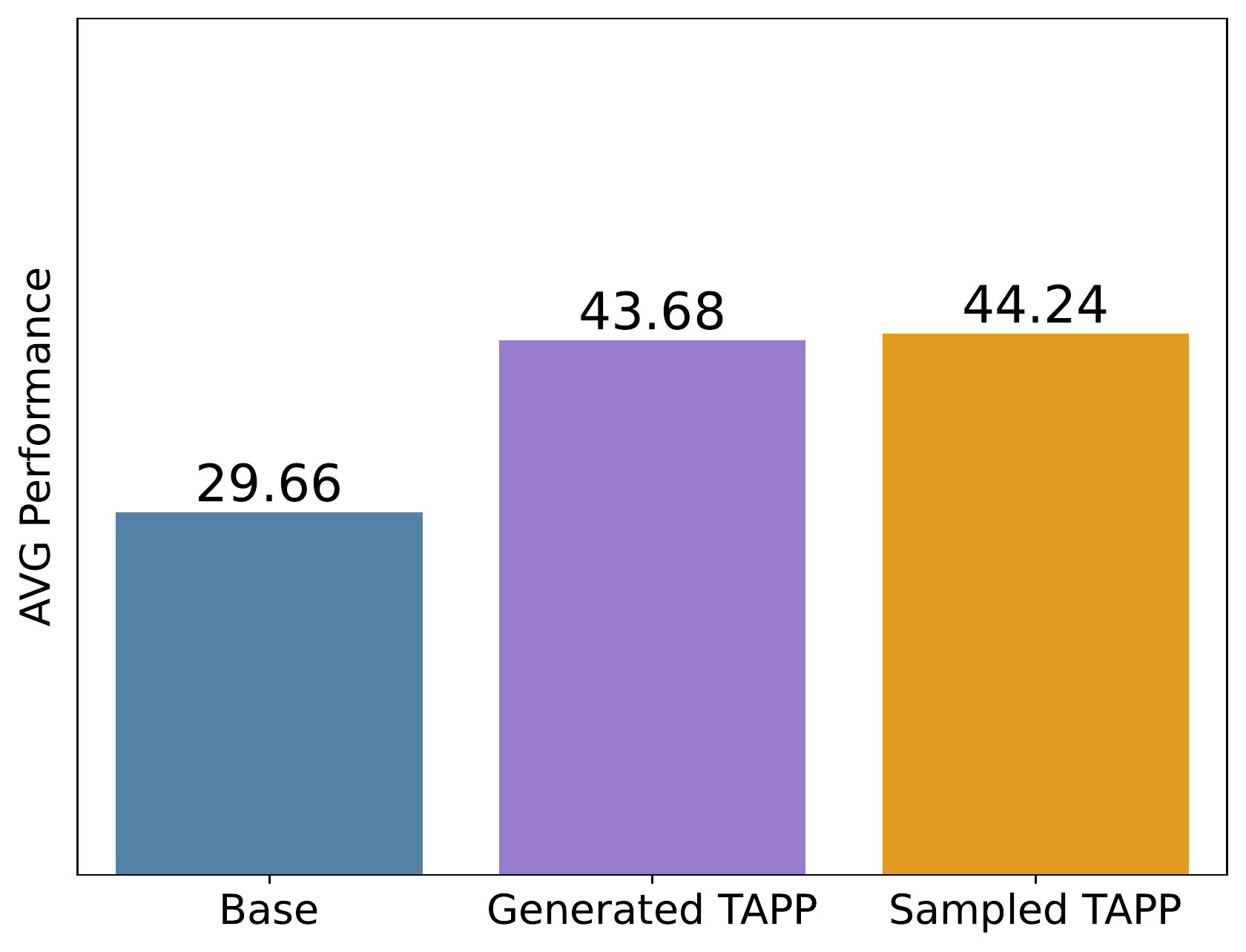}

\caption{The result of \ICIL using demonstrations generated by ChatGPT \citep{chatgpt}. Machine-generated demonstrations show comparable performance to demonstrations sampled from \textsc{SuperNI} benchmark.}
\vspace{-2mm}
\label{fig:chatgpt}
\end{figure}
\begin{figure*}[ht!]
\centering
    \begin{subfigure}[b]{0.3\textwidth}
    \includegraphics[width=\textwidth]{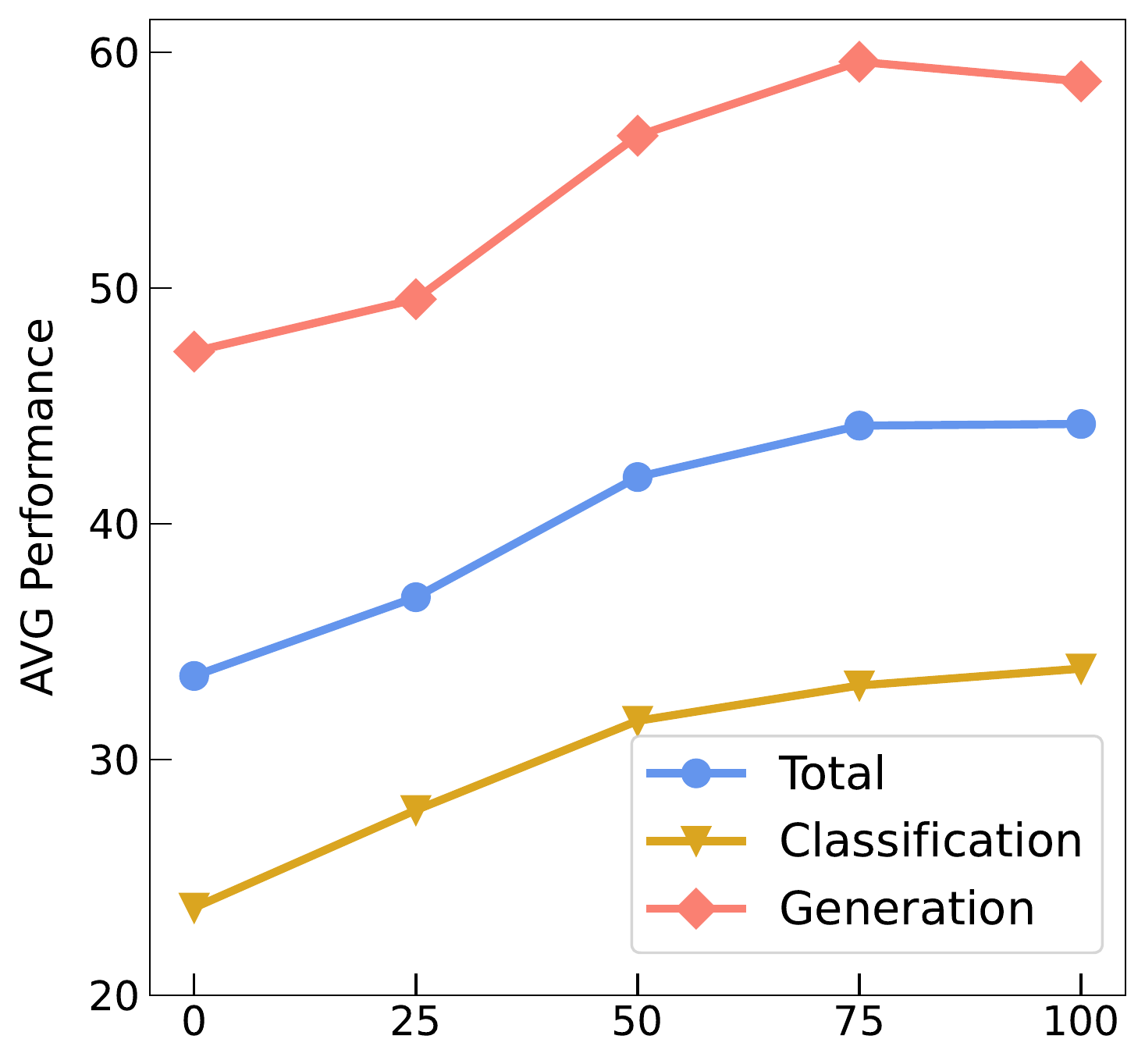}
    \caption{Classification Ratio (\%)}
    \label{fig:classification}
    \end{subfigure}
    \hspace{3mm}
    \begin{subfigure}[b]{0.3\textwidth}
    \includegraphics[width=\textwidth]{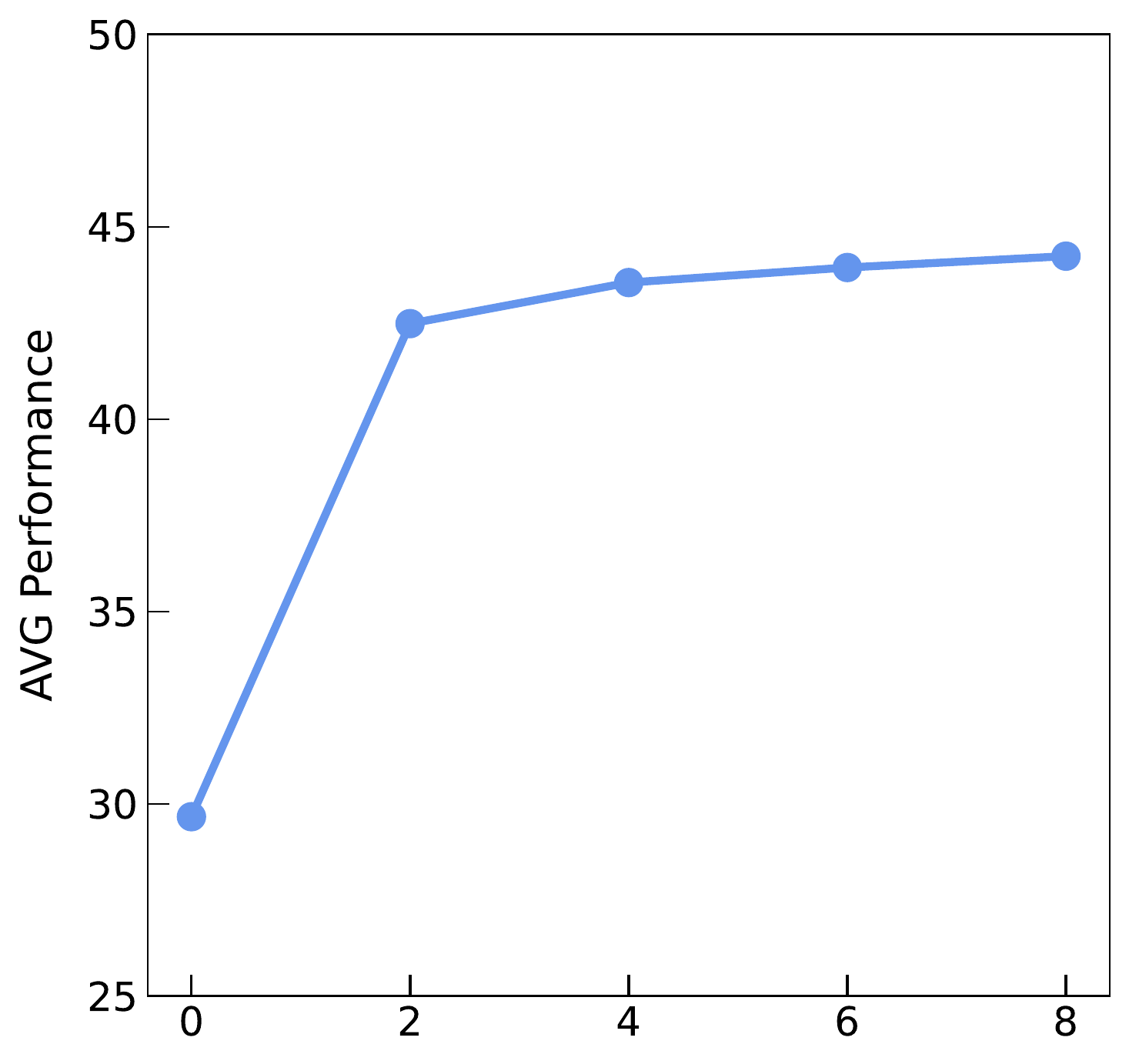}
    \caption{Number of Demonstrations}
    \label{fig:kshot}
    \end{subfigure}
    \hspace{3mm}
    \begin{subfigure}[b]{0.3\textwidth}
    \includegraphics[width=\textwidth]{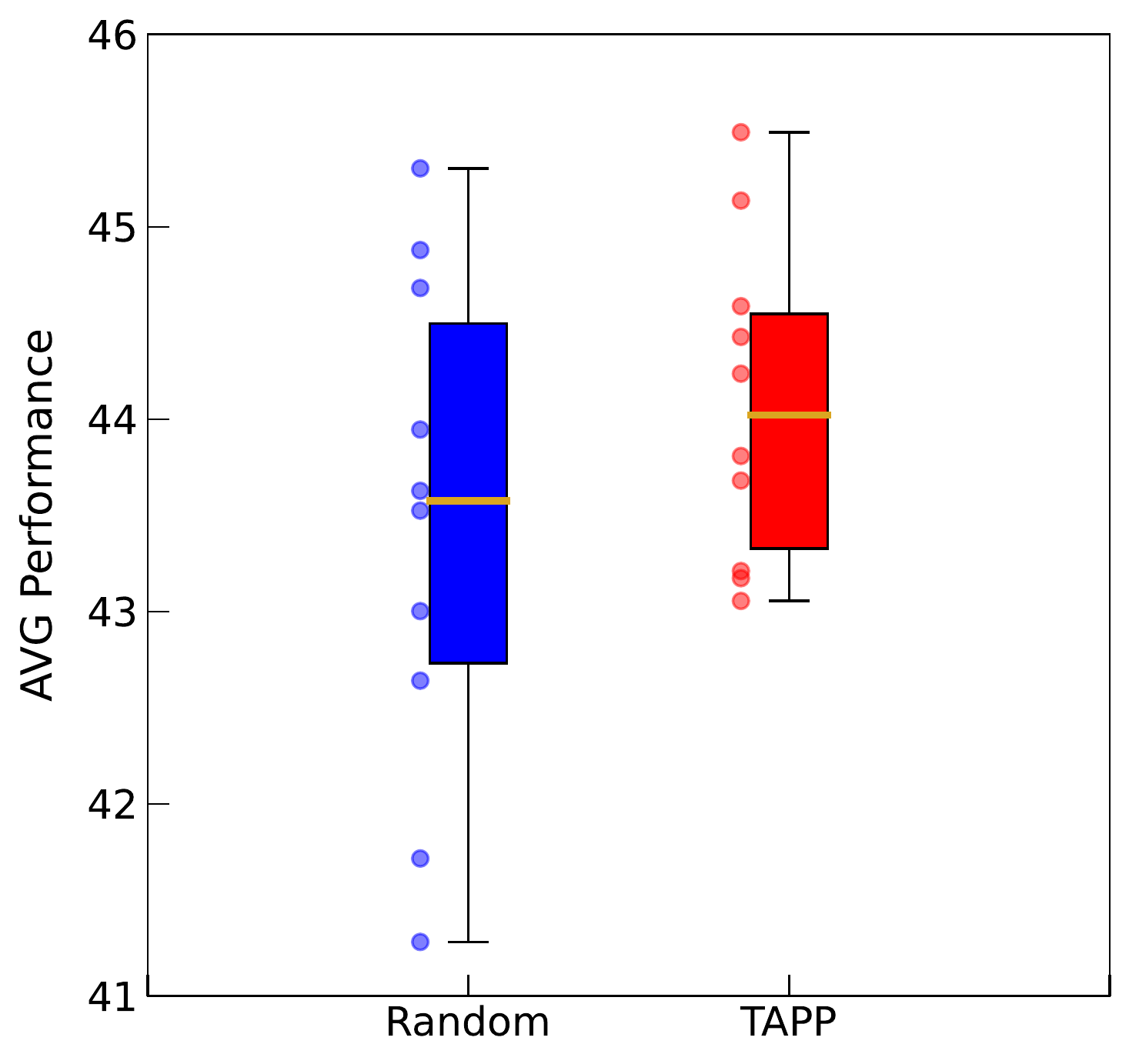}
    \caption{Ordering of Demonstrations}
    \label{fig:variance}
    \end{subfigure}
\caption{(a) shows that the average performance increases as the ratio of classification tasks that are used as demonstrations for \ICIL increases, even for generation target tasks. (b) shows that the performance increases as the number of demonstrations increases for \ICIL. (c) shows that ordering the demonstration set by the number of answer choices reduces the variance on 10 demonstration sets.}
\end{figure*}
We explore if \ICIL shows effectiveness for machine-generated demonstrations instead of sampling from the task pool (\textsc{SuperNI}). We use ChatGPT \citep{chatgpt} for demonstration generation by specifying the rules used to construct the demonstration set for \ICIL. As shown in Figure \ref{fig:chatgpt}, \ICIL is also effective for machine-generated demonstrations, showing comparable performance to \ICIL with demonstrations from \textsc{SuperNI} and significantly outperforming the result without any prefix prompt. This finding suggests that \ICIL is effective even without a sampling process from benchmarks that consist of diverse instructions, indicating that the performance enhancement is not from demonstration construction through sampling, but is from the construction rule and the format of \ICIL. We provide an example of a machine-generated demonstration set in Appendix \ref{appen:chatgpt}.

\subsection{Ablation Studies}
\paragraph{Instruction and output distribution of the demonstrations of matters.}

\begin{table}[]
\centering
\vspace{-2mm}
\fontsize{9}{12}\selectfont
\begin{tabular}{l|ccc|c}
 & Inst. & Input & Output & AVG \\ \midrule
\ICIL & \cmark & \cmark & \cmark & \textbf{44.24} \\
Random Inst. & \xmark & \cmark & \cmark & 31.18 \\
Random Input & \cmark & \xmark & \cmark & \textbf{44.27} \\
Random Output & \cmark & \cmark & \xmark & 38.30 \\
\end{tabular}
\caption{Corrupting the distribution of each component (instruction, input, output) of the demonstration of \ICIL by replacing it with random words or sentences. An example of each demonstration corruption is shown in Table \ref{table:random_example} in the Appendix.}
\label{table:corrupt}
\vspace{-5mm}
\end{table}

We further analyze the effectiveness of each component of the demonstrations for \ICIL by corrupting the distribution of each component: instruction, input, and output instance. For instruction corruption, we replace the ground truth sequences with random sequences from an external corpus, which is similar to how we corrupt the input distribution discussed in Section \ref{sec:results}. For output corruption, we replace ground truth labels with random English words, following \citet{min2022rethinking}. The results are shown in Table \ref{table:corrupt}. Unlike input distribution corruption results of Figure \ref{fig:i3c}, corrupting the distribution of the instruction or the output instance of each demonstration significantly harms the performance. In particular, corrupting the instruction distribution shows little improvement compared to base input without any prefix prompts (31.18 vs 29.67). This suggests that, unlike input instances, the distribution of instruction and output instances significantly affects the performance of \ICIL.
\paragraph{Constructing the demonstration set with classification tasks is important.}

We analyze the heuristic of constructing the demonstration set from only classification tasks in \ICIL by varying the ratio of classification tasks consisting of the demonstration set. As shown in Figure \ref{fig:classification}, the average performance increases as the ratio of classification tasks increases. Interestingly, we observe that constructing the demonstration set with classification tasks also enhances generation (non-classification) target tasks. 
This finding contrasts with few-shot in-context learning setting, where retrieving demonstrations similar to the target query enhances the few-shot performance \citep{rubin2021learning, liu2021gpt} \footnote{Note that the classification ratio of 0\% in Figure \ref{fig:classification} corresponds to constructing the demonstration set solely from generation (non-classification) tasks.}.

\paragraph{Increasing the number of demonstrations improves the performance.}

We study the impact of the number of demonstrations that consist \ICIL. Results are shown in Figure \ref{fig:kshot}. As expected, the mean performance improves as the number of demonstrations increases. Notably, the instruction-following ability significantly improves even with 2 demonstrations, implying that using only a small set of prefix prompts can still improve the performance of LLMs. This also suggests that for settings where efficient computation during inference is crucial, reducing the number of demonstrations that consist \ICIL might be an optimal approach since the performance degradation is not severe.

\paragraph{Ordering the demonstrations by the number of answer choices reduces the variance.}
To examine the impact of different orderings of the demonstration set, we compare the ordering of demonstrations that consist \ICIL based on the number of answer choices with a random ordering. Figure \ref{fig:variance} shows the result of 10 different demonstration sets by sampling them with 10 different random seeds. Although the mean performance does not show a significant difference between the two settings, we observe that applying ordering based on the number of answer choices reduces the variance and improves the worst-case accuracy.

\paragraph{Answer choice overlap between demonstrations harms the performance.}

\begin{table}[]
\centering
\fontsize{9}{11}\selectfont
\begin{tabular}{l|ccc}
\rowcolor[HTML]{FFFFFF} 
 & Classification & Generation & Total \\ \midrule
Overlap & \textbf{35.14} & 52.32 & 42.30 \\
No Overlap & 33.86 & \textbf{58.77} & \textbf{44.24}
\end{tabular}
\caption{Effect of answer choice overlap between demonstrations. The demonstration set that has an overlap underperforms the set without overlap on average, especially for generation tasks.}
\label{table:overlap}
\vspace{-5mm}
\end{table}
We analyze the effect of answer choice overlap between demonstrations, which is one of the rules used to construct the demonstration set. We compare the demonstration set used for \ICIL with the demonstration set that has the same answer choice for all demonstrations. The result is demonstrated in Table \ref{table:overlap}. We observe that the demonstration set with answer choice overlap underperforms the demonstration set without overlap on average, especially for generation tasks. We find that the demonstration set with answer choice overlap tends to make the model generate short sequences for long text generation or predict the output by copying one of the labels of the demonstration set, leading to poor performance.

\section{Discussion}
\label{sec:discussion}
From previous sections, we have observed that \ICIL significantly boosts the performance of both base and instruction-fine-tuned LLMs. Also, we have demonstrated that corrupting the input distribution does not harm the performance much and analyzed that constructing the demonstration set from classification tasks is crucial for performance improvement. In this section, we suggest the role of \ICIL based on the findings from the previous sections. 

\paragraph{Why is constructing the demonstration set from classification tasks important?}
Figure \ref{fig:classification} shows that constructing the demonstration set with classification tasks is important for \ICIL. Then, what is the difference between classification and generation (non-classification) tasks? Because one of our heuristics for demonstration construction is to only consider classification tasks that include an answer choice in the instruction (e.g. \textit{``agent" or ``customer"} in Figure \ref{fig:figure_1}), these demonstrations have more \textit{explicit} cues about the output distribution. We hypothesize that during inference, LLMs learn the correspondence between answer choice in the instruction (e.g. Determine the speaker of the dialogue, ``agent" or ``customer".) and the label (e.g. agent) from demonstrations. Especially, because the label word appears in the instruction for classification tasks, it would be easy to exploit this relationship for LLMs. We observe that adding only a sentence that includes answer choices for corrupted instruction demonstrations in Table \ref{table:corrupt} leads to an increase in the performance of \ICIL ($31.18 \rightarrow 38.92$), supporting the hypothesis. 
\paragraph{What does the result of input-corrupted \ICIL imply?}
From Figure \ref{fig:i3c} and Table \ref{table:corrupt}, we observe that the input distribution of demonstrations for \ICIL does not matter much, while instruction and output distribution matter significantly. This observation bolsters the above hypothesis that LLMs learn the correspondence of answer choice in the instruction and the label of the demonstrations during \ICIL. Instead of relying on complex correspondence such as the relationship between instruction, input, and output altogether, LLMs tend to focus on simple correspondence such as string matching between the instruction including answer choices and the label. Previous work also demonstrates similar findings that LLMs \textit{takes less effort} to adapt to a task, similar to shortcut learning \citep{webson2021prompt, min2022rethinking}.

\paragraph{What is the role of \ICIL?}
If LLMs learn the correspondence of the answer choice in the instruction and the label of the demonstrations during \ICIL, then how does this assist the instruction-following ability? During \ICIL, we hypothesize that the demonstrations give a signal that assists LLMs \textit{focus} on the instruction to more accurately estimate the output distribution, making LLMs better follow instructions. We suggest that this hypothesis explains why constructing the demonstration set from classification tasks also improves the performance of generation target tasks. As a preliminary experiment, we calculate the ratio of how much the first output token attends to the target task instruction compared to the target task input and observe that \ICIL leads to increased attention to the task instruction in Appendix \ref{appen:attention}. Meanwhile, our observations imply that such ability does not seem to be sufficiently activated for both base LLMs and instruction-fine-tuned LLMs. Although instruction fine-tuning also assists the signal of focusing on the instructions, we hypothesize that \ICIL directly enforces the correspondence between the instruction and the label of the demonstrations during inference.

\section{Limitations}
First, although \ICIL leads to significant performance gain across various LLMs, it suffers from increased computation during inference due to the increased number of input sequences. However, since we use a fixed prefix prompt for all tasks, the inference cost can be minimized through caching. Second, our evaluation is mainly based on heuristic metrics such as ROUGE and Exact Match scores. We leave investigating the effect of \ICIL using qualitative evaluation settings by recruiting human evaluators as future work. Third, our interpretation of the role of \ICIL is hypothetical, whereas further interpretation of the role of \ICIL can be conducted by analyzing the inner operation inside the model \citep{lieberum2023does,grosse2023studying}.

\section{Conclusion}
In this paper, we explore the effectiveness of Task-Agnostic Prefix Prompt (\ICIL) for instruction-following during inference of LLMs. We observe that prepending \ICIL that is determined through simple heuristics significantly enhances the performance of both base and instruction-fine-tuned LLMs. \ICIL differs from task-specific prompts in that it is a fixed prompt that can be prepended to any target task. Through detailed analysis, we hypothesize that the effect of \ICIL comes from learning the correspondence between answer choice in the instruction and the label of the classification task demonstrations consisting of \ICIL, leading LLMs to better focus on the instruction. To this end, our work demonstrates the effect of task-agnostic prefix prompts for a diverse set of tasks and suggests research direction for exploring various approaches that further activate the instruction-following ability of LLMs.

\section*{Acknowledgments}
This work was partly supported by Institute of Information \& communications Technology Planning \& Evaluation (IITP) grant funded by the Korea government (MSIT) (No.2022-0-00113, Developing a Sustainable Collaborative Multi-modal Lifelong Learning Framework, 80\%; No.2019-0-00075, Artificial Intelligence Graduate School Program (KAIST), 20\%).
\bibliography{aaai24}

\clearpage
\newpage
\appendix
\section*{Appendix}

\section{Full Results for Each Task Category}
\label{appen:full_result}
On Table \ref{table:pretrained_all} and Table \ref{table:inst_all}, we report full results of various models on \textsc{Super-NaturalInstructions} consisting of 12 categories, specifically the relative performance gain of \ICIL over the standard zero-shot setting. Each task name is shown in abbreviation: 

\begin{multicols}{2}
\begin{itemize}
\item TE: Textual Entailment
\item CEC: Cause Effect Classification
\item CR: Coreference Resolution
\item DAR: Dialogue Act Recognition
\item AC: Answerability Classification
\item WA: Word Analogy
\item OE: Overlap Extraction
\item KT: Keyword Tagging
\item QR: Question Rewriting
\item TG: Title Generation
\item DT: Data to Text
\item GEC: Grammar Error Correction
\end{itemize}
\end{multicols}


\begin{table*}[]
\centering
\fontsize{10}{12}\selectfont
\caption{Relative performance gain achieved by \ICIL over standard zero-shot setting for each task category of \textsc{SuperNI} benchmark on various pretrained LLMs. We observe that the number of tasks that benefit from \ICIL increases as the model size scales up.}
\begin{tabular}{lcccccccc}
\midrule
\multirowcell{2}{}         & \textbf{OPT}          & \textbf{Curie}         & \textbf{GPT-J}        & \textbf{OPT}            & \textbf{NeoX}  & \textbf{OPT} & \textbf{OPT}  & \textbf{Davinci}        \\
      & \textbf{6.7B}             & \textbf{6.7B}             & \textbf{6B}   & \textbf{13B}    & \textbf{20B}   & \textbf{30B}               & \textbf{60B} & \textbf{175B}             \\ \midrule
TE           & 27.15                                                                              & 18.11                                                                                & 5.31                                                                               & 29.50                                                                             & 13.93                                                                                & 22.10                                                                             & 2.93                                                                             & 2.85                                                                                   \\
CEC  & 22.00                                                                              & 24.62                                                                                & 4.90                                                                               & 27.90                                                                             & 8.52                                                                                 & 16.86                                                                             & -0.62                                                                            & 4.43                                                                                   \\
CR      & 12.05                                                                              & 8.07                                                                                 & 10.55                                                                              & 14.57                                                                             & 8.29                                                                                 & 11.52                                                                             & -1.19                                                                            & 13.95                                                                                  \\
DAR     & 23.33                                                                              & 27.57                                                                                & 22.00                                                                              & 27.95                                                                             & 12.57                                                                                & 18.14                                                                             & 1.67                                                                             & 21.14                                                                                  \\
AC & 22.21                                                                              & 18.64                                                                                & 8.33                                                                               & 36.64                                                                             & 16.15                                                                                & 23.36                                                                             & 1.69                                                                             & 5.64                                                                                   \\
WA              & 9.96                                                                               & 12.21                                                                                & 16.92                                                                              & 8.38                                                                              & 21.17                                                                                & 7.29                                                                              & 1.75                                                                             & 20.71                                                                                  \\ 
OE          & 5.26                                                                               & -11.74                                                                               & -9.52                                                                              & -8.14                                                                             & -9.53                                                                                & 4.42                                                                              & -1.94                                                                            & 3.98                                                                                   \\
KT             & 20.14                                                                              & 2.38                                                                                 & 14.83                                                                              & 13.45                                                                             & 6.69                                                                                 & 17.80                                                                             & 0.92                                                                             & 15.96                                                                                  \\
QR         & -24.43                                                                             & -19.56                                                                               & -26.78                                                                             & -18.27                                                                            & -13.33                                                                               & -19.14                                                                            & 0.08                                                                             & 3.06                                                                                   \\
TG           & -0.10                                                                              & -0.59                                                                                & -4.70                                                                              & 9.21                                                                              & 5.63                                                                                 & 6.47                                                                              & -2.21                                                                            & 5.71                                                                                   \\
DTT              & -5.57                                                                              & -8.11                                                                                & -0.96                                                                              & 0.07                                                                              & -1.00                                                                                & 3.63                                                                              & 1.61                                                                             & 1.89                                                                                   \\
GEC     & -18.72                                                                             & -27.59                                                                               & -23.34                                                                             & -3.59                                                                             & -3.95                                                                                & -20.99                                                                            & 0.94                                                                             & 0.84                                                                                   \\ 
\end{tabular}

\label{table:pretrained_all}
\end{table*}


\begin{table*}[]
\centering
\fontsize{10}{12}\selectfont
\caption{Relative performance gain achieved by \ICIL over standard zero-shot setting for each task category of \textsc{SuperNI} benchmark on instruction-fine-tuned LLMs.}
\begin{tabular}{lcccc}
\midrule
\multirowcell{2}{}         & \textbf{Curie-001}          & \textbf{Davi-001}         & \textbf{Davi-002}        & \textbf{Davi-003}            \\
      & \textbf{6.7B}             & \textbf{175B}            & \textbf{175B} & \textbf{175B}             \\ \midrule
TE           & 9.65                                   & 9.19                                     & 7.89                                     & -0.63                                    \\
CEC  & 6.86                                   & 0.29                                     & 9.05                                     & 5.43                                     \\
CR      & 4.38                                   & 13.52                                    & 5.10                                     & 0.67                                     \\
DAR     & 7.29                                   & 15.33                                    & 15.19                                    & 15.29                                    \\
AC & 3.64                                   & 0.51                                     & 2.79                                     & 3.54                                     \\
WA                & 3.33                                   & 16.50                                    & 25.75                                    & 27.92                                    \\ 
OE          & -9.27                                  & 1.17                                     & 3.37                                     & 1.96                                     \\
KT              & -5.54                                  & -3.02                                    & 20.01                                    & 9.62                                     \\
QR           & -12.03                                 & -3.37                                    & 14.54                                    & 2.49                                     \\
TG             & -5.32                                  & 1.54                                     & 5.19                                     & 2.42                                     \\
DTT               & 4.84                                   & 3.96                                     & 3.11                                     & 1.36                                     \\
GEC     & -13.41                                 & 2.02                                     & 8.69                                     & 1.11                                     \\ 
\end{tabular}
\label{table:inst_all}
\end{table*}

\section{Evaluation using INSTRUCTSCORE}
\label{appen:instructscore}
\begin{figure}[ht!]
\centering
\includegraphics[width=0.5\textwidth]{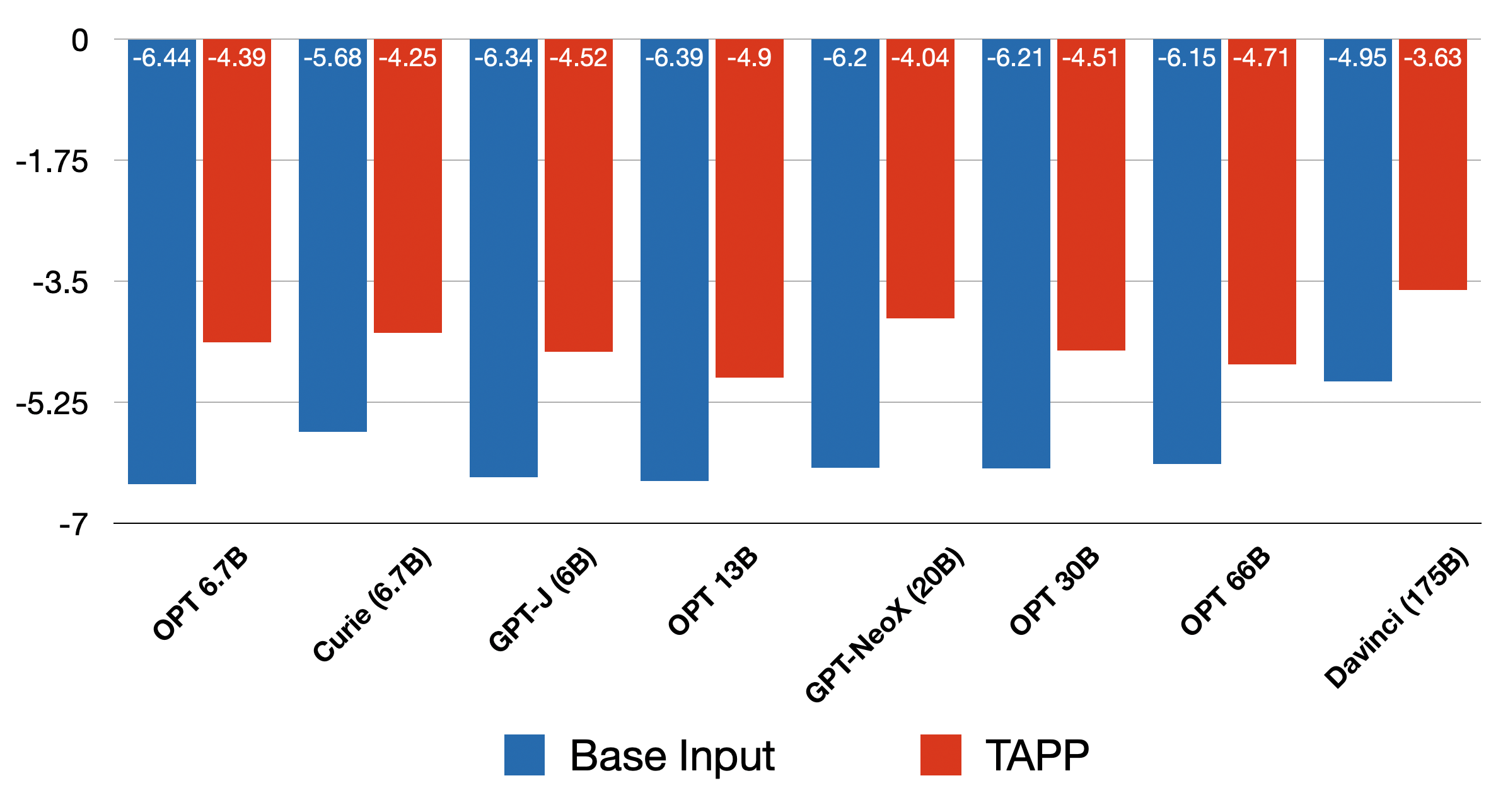}
\caption{Evaluation of various LLMs on 12 tasks using INSTRUCTSCORE metric \citep{xu2023instructscore}. TAPP shows effectiveness across all models.}
\label{fig:instructscore}
\end{figure}

For the main experiments, we have evaluated various models using heuristic metrics such as Exact Match (EM) or ROUGE-L score. However, since these metrics might not fully represent the \textit{instruction-following} ability of LLMs, we also conduct an evaluation using INSTRUCTSCORE \citep{xu2023instructscore}, which uses a fine-tuned 7B LLaMA model \citep{touvron2023llama} to evaluate the response by counting the number of major errors (score of -5) and minor error (score of -1), and measuring the final score based on the number of errors. Due to the computation, we evaluate models on the 12 subset tasks mentioned in Section \ref{sec:ablation}. As shown in Figure \ref{fig:instructscore}, for all models, \ICIL enhances the INSTRUCTSCORE performance. This indicates that TAPP indeed leads to the enhancement of instruction-following ability. 

\section{Preliminary Attention Analysis}
\label{appen:attention}
In Section \ref{sec:ablation}, we hypothesize that the role of \ICIL is to make LLMs focus more on the instruction. As a preliminary experiment to confirm this, we conduct an attention analysis of OPT-6.7B model for 5 generation tasks which especially necessitate comprehension of the instruction and cannot be solved with input copying heuristic baseline, which is a total of 500 instances. To test our hypothesis that `TAPP potentially makes LLMs focus more on the instruction', we measure the ratio of how much the first output token attends to the target task instruction compared to the target task input as defined below:

\begin{equation}
   InstAttn =  \frac{\frac{1}{|I|}\sum_{l=1}^{L} \sum_{h=1}^{H} \sum_{t \in I} a_{lh}(o, t)}{\frac{1}{|X|}\sum_{l=1}^{L} \sum_{h=1}^{H} \sum_{t \in X} a_{lh}(o, t)}
\end{equation}

where $L$, $H$ denote the number of layers and heads of the model, $a_{lh}$ corresponds to the attention value for the $h$th head of the $l$th layer, $o$ corresponds to the first output token, and $I$ and $X$ denote the set of target task instruction and input instance tokens, respectively.
We observe that TAPP increases the value of $InstAttn$ by 10.39\% on average compared to the base input setting (\(0.6172 \rightarrow 0.6758\)), supporting our hypothesis. We leave more rigorous analysis as future work.

\section{Preliminary Observation on OPT-IML and FLAN-T5}

From preliminary experiments, we observe that applying \ICIL on OPT-IML (30B) and FLAN-T5 underperforms the standard zero-shot setting. For OPT-IML, we suggest the degradation is from the characteristics of OPT-IML that few-shot in-context learning underperforms zero-shot setting, especially for OPT-30B. As shown in the results of \citet{iyer2022opt}, increasing the number of few-shot examples harms the held-out evaluation results of OPT-IML (30B) on \textsc{SuperNI} benchmark. This indicates that prepending the demonstration set \textit{distracts} the zero-shot task adaptation. For FLAN-T5, we observe that the degradation of applying \ICIL is due to predicting the output by copying from the demonstration label. This is an undesirable behavior for \ICIL because the target task and the task of the demonstrations are different. We suggest this copying behavior occurs because FLAN-T5 was explicitly trained to do in-context learning. Therefore, the model would interpret the cross-task demonstration set of \ICIL as the target task demonstration for few-shot in-context learning. Therefore, it would lead to the model copying one of the labels from the demonstrations, harming the performance.
\label{appen:opt_iml}
 

\section{Evaluation Setting Details}
\label{appen:evaluation_setting}
For all evaluation settings, we set the stop sequence as "\textbackslash n\textbackslash n". Also for GPT-3 models, we set the maximum input and output sequence length as 2048 and 128 respectively. For other models, we set the maximum input and output sequence length as 1024 and 64 respectively. For target task instances that are long which makes the concatenation of $K$ demonstrations and the target task input sequence exceed the maximum sequence length, we only include the front $K' (K'<K)$ demonstrations that fit the max sequence length. For standard zero-shot setting, we follow the format of \citet{wang2022benchmarking}, appending a sentence "Now complete the following example-" in front of the target input instance. From preliminary experiments, we observe that prepending this sentence improves the performance for standard zero-shot setting. For \ICIL and few-shot in-context learning experiment, we do not include such sentence. 

\section{Qualitative Evaluation}
\begin{figure*}
    \centering
    \includegraphics[width=\linewidth]{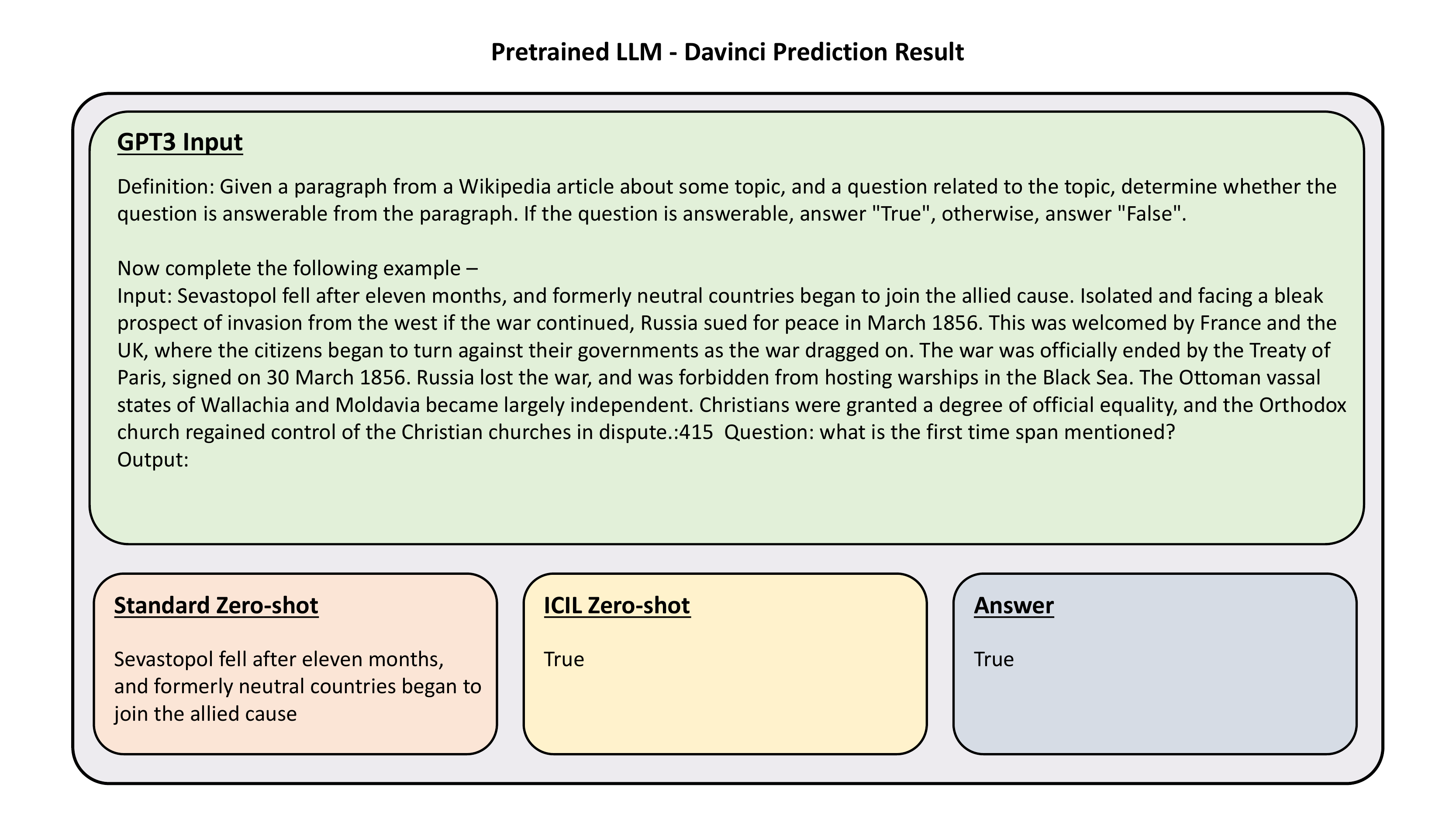}
    \caption{Qualitative Example of responses to one of the evaluation instances from \textsc{SuperNI} benchmark, comparing the responses of standard zero-shot setting and \ICIL of GPT-3 davinci model}
    \label{fig:qual_1}
\end{figure*}

\begin{figure*}
    \centering
    \includegraphics[width=\linewidth]{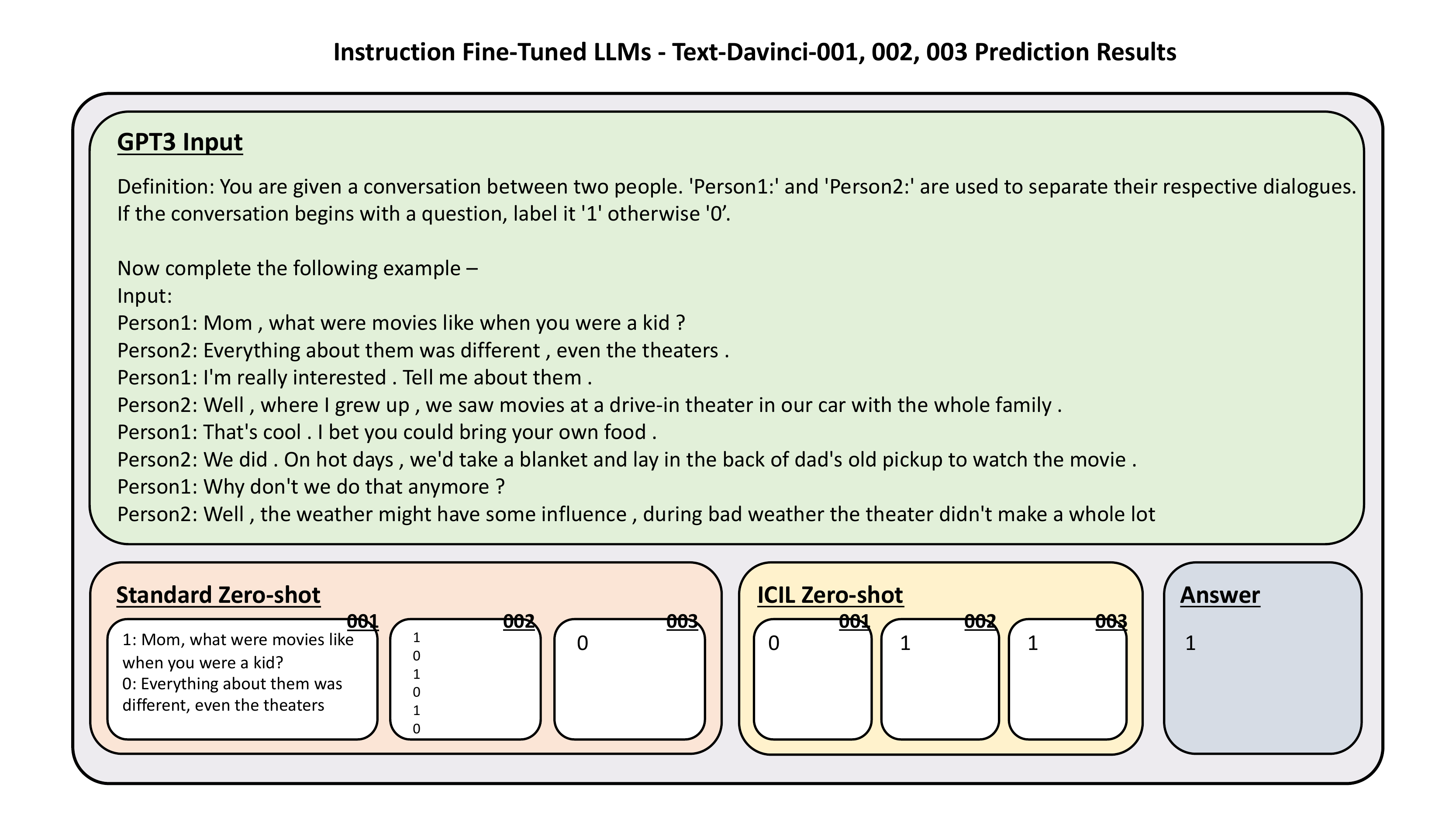}
    \caption{Qualitative Example of responses to one of the evaluation instances from \textsc{SuperNI} benchmark, comparing the responses of standard zero-shot setting and \ICIL of GPT-3 text-davinci-001, 002, 003.}
    \label{fig:qual_2}
\end{figure*}

Figure \ref{fig:qual_1} and Figure \ref{fig:qual_2} shows the examples of cherry-picked examples of responses to evaluation instances from \textsc{SuperNI} benchmark.

\section{Example of Model-Generated Prompts}
\label{appen:chatgpt}
Figure \ref{fig:chatgpt_demo} shows the list of demonstrations that are generated by ChatGPT. We manually added/revised some parts that the model did not follow the heuristics such as not including the answer choices in the instruction.

\begin{figure*}
\begin{tcolorbox}
\begin{scriptsize} 
Definition: In this task, you will be performing image classification on an image of a bird. You have to select the correct species of the bird from the options provided: "Pigeon" or "House Sparrow" \newline

Input: A picture of a small bird with brown and white feathers sitting on a tree branch. \newline
Output: House Sparrow \newline

Definition: In this task, you will be identifying named entities from a given text. You have to identify the organization name mentioned in the following news article. Choose from 'Apple' or 'Google'. \newline

Input: The CEO of Google, Sundar Pichai, announced the launch of the company's latest project in collaboration with NASA. \newline
Output: Google \newline

Definition: In this task, you will be performing text classification on a social media post. You have to classify the post into one of the following categories: personal, professional, or social. \newline

Input: Just landed in Paris for my dream vacation! Can't wait to explore the city of love! \#paris\#vacation\#travel \newline
Output: personal \newline

Definition: In this task, you will be performing text classification on a product review. You have to classify the review into one of the following categories: usability, performance, or design. \newline

Input: The new laptop has a sleek and modern design. The keyboard is easy to use and the touchpad is very responsive. However, the battery life is not as good as expected. \newline
Output: design \newline

Definition: In this task, you will be performing text classification on a news article. You have to classify the article into one of the following categories: politics, sports, or entertainment. \newline

Input: The Indian government has proposed a new budget for the upcoming financial year. The budget focuses on healthcare and infrastructure development, and aims to boost the country's economic growth. The opposition parties have criticized the budget, claiming that it neglects the needs of the common people. \newline
Output: politics \newline

Definition: In this task, you will be performing sentiment analysis on a customer review. You have to identify the sentiment of the review as either positive, negative or neutral. Read the following customer review and select the sentiment from the options provided. \newline

Input: I recently purchased this product and I must say I am extremely happy with it. The quality is exceptional and it has exceeded my expectations. I would highly recommend this product to anyone looking for a reliable and durable option. \newline
Output: positive \newline

Definition: In this task, you will be performing speech emotion recognition on an audio clip. You have to identify the emotion expressed in the audio clip as either happy, sad, angry, or neutral. \newline

Input: An audio clip of a person saying, I am so excited to be going on vacation next week! \newline
Output: happy \newline
 
Definition: In this task, you will be performing image classification on an image of a dog. You have to select the correct breed of the dog from the options provided. Options: 'Chihuahua', 'Poodle', 'Bulldog', 'Border Collie', 'Golden Retriever'. \newline

Input: A picture of a medium-sized dog with short brown fur, droopy ears, and a wrinkled face. \newline
Output: Bulldog
\end{scriptsize}
\end{tcolorbox}
\caption{Example of model-generated demonstration set.}
\label{fig:chatgpt_demo}
\end{figure*}

\section{List of Prompts for \ICIL}
\label{appen:prom_i2c}

In Figure \ref{fig:i2c_seed1}, Figure \ref{fig:i2c_seed2}, and Figure \ref{fig:i2c_seed3}, we list out the fixed prompts (demonstration set) that are used for the evaluation of \ICIL.

\begin{figure*}
\begin{tcolorbox}
\begin{scriptsize} 
Definition: In this task, you are given a dialogue from a conversation between an agent and a customer. Your task is to determine the speaker of the dialogue. Answer with "agent" or "customer". \newline

Input: I have successfully booked your ticket with flight-1017, have a safe journey. \newline 
Output: agent \newline

Definition: In this task, you are given a hateful post in English from online platforms. You are expected to classify the post into two classes: aggresive or non-aggresive. An aggresive post is one that expresses feelings and opinions in an abusive way and tries to dominate others. Note that the URLs in the text have been replaced with [Link]. \newline

Input: Post: @burnthewitchiii come tell this bitch to get in the fucking sea. Please. Before I cunt punch her. \newline
Output: Aggresive \newline

Definition: In this task you are given a medical question pair hand-generated. Your task is to classify a given post into two categories 1) 'Similar' if the given two questions have a same connotation 2) 'Dissimilar' if the given two questions have a different connotation or meaning. \newline

Input: Sentence1: Should I ask to be put on a Statin at my checkup tomorrow? My cholesterol was very high and I'm sure it still is. 
 Sentence2: What are the side effects of statin medications? \newline
Output: Dissimilar \newline

Definition: We would like you to assess the QUALITY of each of the following argument (discussing Gun Control) and determine if the argument is Valid or Invalid. A valid argument is clearly interpretable and either expresses an argument, or a premise or a conclusion that can be used in an argument for the topic of gun control. An invalid argument is a phrase that cannot be interpreted as an argument or not on the topic of gun control. \newline

Input: i think that SOCOM would love guns arms that cant be dected, it would alco be useful for undercover agents. \newline
Output: Valid \newline

Definition: You will be given two pieces of text with the same meaning. One of them is simpler and easier to understand for non-native English speakers. Complex texts may contain more difficult words, have unnecessary phrases or contain long sentences. Your task is to choose the simpler piece of text. You are expected to output 'Text one' if the first sentence is simpler. Otherwise output 'Text two'. \newline

Input: Text one: Rollins retired in 1962 and opted to become a coach.
 Text two: Rollins retired in 1962. He then chose to become a coach. \newline
Output: Text two \newline

Definition: You are given a target person and two sentences. Your task is to determine whether the second sentence conveys either a stereotype, anti-stereotype, or is unrelated with respect to the first sentence and the target.  A stereotype is a widely held but oversimplified notion of a group of people. An anti-stereotype is an idea that goes against the widely held but oversimplified notion of a group of people. Sentences and targets are unrelated if there is no relevance between them. \newline

Input: Target: himself \newline
He is proud of himself. My phone is off. \newline
Output: unrelated \newline

Definition: You are given a question. You need to detect which category better describes the question. A question belongs to the description category if it asks about description and abstract concepts. Entity questions are about entities such as animals, colors, sports, etc. Abbreviation questions ask about abbreviations and expressions abbreviated. Questions regarding human beings, description of a person, and a group or organization of persons are categorized as Human. Quantity questions are asking about numeric values and Location questions ask about locations, cities, and countries. Answer with "Description", "Entity", "Abbreviation", "Person", "Quantity", and "Location". \newline

Input: What is the oldest building in the United States ? \newline
Output: Location \newline

Definition: You will be given two sentences. One of them is created by paraphrasing the original one, with changes on an aspect, or using synonyms. Your task is to decide what is the difference between two sentences. Types of change are explained below: \newline
 Tense: The verbs in the sentence are changed in tense. \newline
Number: Plural nouns, verbs and pronouns are changed into single ones or the other way around. \newline
Voice: If the verbs are in active voice, they're changed to passive or the other way around. \newline
Adverb: The paraphrase has one adverb or more than the original sentence. \newline
Gender: The paraphrase differs from the original sentence in the gender of the names and pronouns. \newline
Synonym: Some words or phrases of the original sentence are replaced with synonym words or phrases. Changes in the names of people are also considered a synonym change. Classify your answers into Tense, Number, Voice, Adverb, Gender, and Synonym. \newline

Input: original sentence: Jim yelled at Kevin because he was so upset . \newline paraphrase: Jim violently yelled at Kevin because he was so upset . \newline
Output: Adverb
\end{scriptsize}
\end{tcolorbox}
\caption{Fixed prompt (Demonstration set) for evaluation of \ICIL, Example 1}
\label{fig:i2c_seed1}
\end{figure*}


\begin{figure*}
\begin{scriptsize}
\begin{tcolorbox}
Definition: In this task, you are given the name of an Indian food dish. You need to classify the dish as "sweet" or "spicy". \newline

Input: Dharwad pedha \newline
Output: sweet  \newline

Definition: In this task, you are given two natural language statements with similar wording. You must choose the statement that makes less sense based on common sense knowledge. A '
' separates the statements. Use "first" or "second" to indicate which sentence makes less sense.  \newline

Input: He played the cow very well  \newline
He played the harp very well  \newline
Output: first  \newline

Definition: In this task, you are given a public comment from online platforms. You are expected to classify the comment into two classes: sexual-explicit and non-sexual-explicit. A comment is considered sexual-explicit if it explicitly portrays sexual matters.  \newline

Input: Comment: Actually, being wiretapped isn't something they have in common.  Merkel was tapped, Trump wasn't.  The man has more than a few screws loose.  \newline
Output: Non-sexual-explicit  \newline

Definition: In this task, you are given a sentence. You must judge whether a single noun or verb has been replaced with another word with the same part of speech. The inversion would result in the sentence sounding unnatural, So unnatural sentences will be considered changed. Label the instances as "Original" or "Changed" based on your judgment.  \newline

Input: Computers are just systems with a great amount of unconsciousness : everything held in immediate memory and subject to programs which the operator initiates .  \newline
Output: Original  \newline

Definition: You are given a short poem which is written by a kid. Your task is to predict if the kid is in elementary school or high school. Anyone who is from grade 1st to 6th-grade will be considered as in elementary school, and 7th to 12th-grade kids will be considered as in high school. There are only two possible outputs, i.e., elementary and high. All inputs have at least one output and it cannot have both categories at a time.  \newline

Input: pink be what i see skateboard be fun go home be what i want to see but when i sleep my day be do  \newline
Output: elementary  \newline

Definition: In this task, you are given a text from a social media post. Your task is to classify the given post into two categories: 1) yes if the given post is intentionally offensive, 2) no, otherwise. Also, generate label 'no' for offensive statements that appear to be unintentional, or non-offensive statements that are underhandedly offensive.  Warning: the examples and instances may contain offensive language.  \newline

Input: RT @CreationOfJay: No girl sucks dick but somehow all dudes get head. Dudes only fuck bad bitches yet all these ugly girls getting pregnant\&\#8230;  \newline
Output: Yes  \newline

Definition: In this task, you are given a public comment from online platforms. You are expected to classify the comment into two classes: toxic and non-toxic. Toxicity is defiend as anything that is rude, disrespectful, or unreasonable that would make someone want to leave a converation.  \newline

Input: Comment: MAGA! MAGA! MAGA! MAGA! MAGA! MAGA! MAGA! MAGA! MAGA! MAGA! MAGA! MMAGA! MAGA! MAGA! MAGA! MAGA! MAGA! MAGA! MAGA! MAGA! MAGA! MAGA! MAGA! MAGA! MAGA! MAGA! MAGA! MMAGA! MAGA! MAGA! MAGA! MAGA! MAGA! MAGA! MAGA! MAGA! MAGA! MAGA! MAGA! MAGA! MAGA! MAGA! MAGA! MMAGA! MAGA! MAGA! MAGA! MAGA!  \newline
Output: Non-toxic  \newline

Definition: Given a comment text in Malayalam, classify the comment into one of these categories (i) Hope speech, (ii) Not Hope Speech or (iii) Not in Expected Language. A hope speech contains content that is encouraging, positive or supportive contents and talks about equality, diversity or inclusion  \newline

Input: avare njaan kutta pedittilla society  oru kaaranama baaki njan taazhe commente cheythattond  \newline
Output: Hope Speech
\end{tcolorbox}
\end{scriptsize}

\caption{Fixed prompt (Demonstration set) for evaluation of \ICIL, Example 2}
\label{fig:i2c_seed2}
\end{figure*}

\begin{figure*}
\begin{scriptsize}
\begin{tcolorbox}
Definition: You will be given a topic and an argument. Decide the argument's stance towards that topic. The argument's stance is in favor or against the topic. If the argument supports that topic, answer with "in favor"; otherwise, if the argument opposes the topic, answer with "against".  \newline

Input: topic: New START Treaty  \newline
argument: Delay risks dangerous non-ratification.  \newline
Output: in favor  \newline

Definition: You are given an array of integers, check if it is monotonic or not. If the array is monotonic, then return 1, else return 2. An array is monotonic if it is either monotonically increasing or monotonocally decreasing. An array is monotonically increasing/decreasing if its elements increase/decrease as we move from left to right  \newline

Input: [6, 12, 18, 24, 30, 36, 42, 48, 54, 60, 66, 72, 78, 84, 90, 96, 102, 108]  \newline
Output: 1  \newline

Definition: Given a sentence, judge the quality of this sentence by indicating "Good" and "Bad". The quality depends on the grammar and the meaning of the sentence. If a sentence is easily understandable, and doesn't have grammatical errors, answer with "Good", otherwise answer with "Bad".  \newline

Input: But a 1978 article by Ted Bear, then a historian at Edward Air Force Base where the alleged experiment took place states  \newline
Output: Good  \newline

Definition: We would like you to assess the QUALITY of each of the following argument (discussing Gun Control) and determine if the argument is Valid or Invalid. A valid argument is clearly interpretable and either expresses an argument, or a premise or a conclusion that can be used in an argument for the topic of gun control. An invalid argument is a phrase that cannot be interpreted as an argument or not on the topic of gun control.  \newline

Input: I posted the real story of the Ft. Hood incident and also a quote that only a good man with a gun can stop a bad man with a gun.  \newline
Output: Valid  \newline

Definition: Classify the given tweet into the three categories: (1) 'Hate Speech', (2) 'Offensive' and (3) 'Neither'. 'Hate Speech' is kind of a threating statement or sometimes include call for violence while 'offensive' statement just offensds someone. 'Neither' is when it doesn't fall into Hate Speech or Offensive category.  \newline

Input: @gonzalez\_sassy ur not sassy white trash change ur username  \newline
Output: Hate Speech  \newline

Definition: In this task, you are given a news article. Your task is to classify the article to one out of the four topics 'World', 'Sports', 'Business', 'Sci/Tech' if the article's main topic is relevant to the world, sports, business, and science/technology, correspondingly. If you are not sure about the topic, choose the closest option. Note that URLs in the text have been replaced with [Link].  \newline

Input: Bone Loss a Serious Threat to Older Americans By LAURAN NEERGAARD    WASHINGTON (AP) -- Half of Americans older than 50 will be at risk of fractures from too-thin bones by 2020, the surgeon general warned Thursday, urging people to get more calcium, vitamin D and exercise to avoid crippling osteoporosis.    The bone-thinning disease is on the rise as the population grays - but weak bones aren't a natural consequence of aging, Surgeon General Richard Carmona stressed...
Output: Sci/Tech  \newline

Definition: In this task, you are given Twitter posts. Your task is to label the post's emotion (as expressed by the user) as sadness, joy, love, anger, fear, or surprise.  \newline

Input: i can tell you the things i don t feel that maybe i should be feeling but i can t really put my finger on the cause of my being shaken  \newline
Output: fear  \newline

Definition: You are given a question. You need to detect which category better describes the question. A question belongs to the description category if it asks about description and abstract concepts. Entity questions are about entities such as animals, colors, sports, etc. Abbreviation questions ask about abbreviations and expressions abbreviated. Questions regarding human beings, description of a person, and a group or organization of persons are categorized as Human. Quantity questions are asking about numeric values and Location questions ask about locations, cities, and countries. Answer with "Description", "Entity", "Abbreviation", "Person", "Quantity", and "Location".  \newline

Input: Who is the current prime minister and president of Russia ?  \newline
Output: Person
\end{tcolorbox}
\end{scriptsize}
\caption{Fixed prompt (Demonstration set) for evaluation of \ICIL, Example 3}
\label{fig:i2c_seed3}
\end{figure*}


\section{List of Prompts for Input-corrupted \ICIL}
\label{appen:i3c}
We list out the fixed prompts (demonstration sets) that are used for evaluation of Input-corrupted \ICIL in Figure \ref{fig:i3c_seed1}, Figure \ref{fig:i3c_seed2}, and Figure \ref{fig:i3c_seed3}, which randomly replace input sentences of each demonstration set shown in Appendix \ref{appen:prom_i2c}.

\begin{figure*}
\begin{scriptsize} \begin{tcolorbox}
Definition: In this task, you are given a dialogue from a conversation between an agent and a customer. Your task is to determine the speaker of the dialogue. Answer with "agent" or "customer". \newline

Input: They were taken the Beaufort County Detention Center and given a \$100,000 bond. \newline
Output: agent \newline

Definition: In this task, you are given a hateful post in English from online platforms. You are expected to classify the post into two classes: aggresive or non-aggresive. An aggresive post is one that expresses feelings and opinions in an abusive way and tries to dominate others. Note that the URLs in the text have been replaced with [Link]. \newline

Input: The market research analysis has been structured using vital data from industry expertize. WhatsApp Or is it a key-tar?  \newline
Output: Aggresive \newline

Definition: In this task you are given a medical question pair hand-generated. Your task is to classify a given post into two categories 1) 'Similar' if the given two questions have a same connotation 2) 'Dissimilar' if the given two questions have a different connotation or meaning. \newline

Input: Learn more. The prosecutors said Wall died either by strangulation or having her throat cut. In other more populated areas it will fall to around -5C. Contact She keeps the place tidy and recently repainted.  \newline
Output: Dissimilar \newline

Definition: We would like you to assess the QUALITY of each of the following argument (discussing Gun Control) and determine if the argument is Valid or Invalid. A valid argument is clearly interpretable and either expresses an argument, or a premise or a conclusion that can be used in an argument for the topic of gun control. An invalid argument is a phrase that cannot be interpreted as an argument or not on the topic of gun control. \newline

Input: There are many institutes in this country which have specialised in certain areas and have highly qualified faculty for the same.  \newline
Output: Valid \newline

Definition: You will be given two pieces of text with the same meaning. One of them is simpler and easier to understand for non-native English speakers. Complex texts may contain more difficult words, have unnecessary phrases or contain long sentences. Your task is to choose the simpler piece of text. You are expected to output 'Text one' if the first sentence is simpler. Otherwise output 'Text two'. \newline

Input: In some ways, he seemed the classic Christian Democrat, stressing moderation and compromise. Preparing for Hurricane Season\textbackslash n by Andrew James : Yes, not -- don't ask Jesse.  \newline
Output: Text two \newline

Definition: You are given a target person and two sentences. Your task is to determine whether the second sentence conveys either a stereotype, anti-stereotype, or is unrelated with respect to the first sentence and the target.  A stereotype is a widely held but oversimplified notion of a group of people. An anti-stereotype is an idea that goes against the widely held but oversimplified notion of a group of people. Sentences and targets are unrelated if there is no relevance between them. \newline

Input: All rights reserved. So why hasn't it happened? wakefieldwall Strong-running second rower \newline 
Output: unrelated \newline

Definition: You are given a question. You need to detect which category better describes the question. A question belongs to the description category if it asks about description and abstract concepts. Entity questions are about entities such as animals, colors, sports, etc. Abbreviation questions ask about abbreviations and expressions abbreviated. Questions regarding human beings, description of a person, and a group or organization of persons are categorized as Human. Quantity questions are asking about numeric values and Location questions ask about locations, cities, and countries. Answer with "Description", "Entity", "Abbreviation", "Person", "Quantity", and "Location". \newline

Input: Jack Dapore and Jordan York both scored 9 points for Russia. \newline 
Output: Location \newline

Definition: You will be given two sentences. One of them is created by paraphrasing the original one, with changes on an aspect, or using synonyms. Your task is to decide what is the difference between two sentences. Types of change are explained below: \newline
 Tense: The verbs in the sentence are changed in tense. \newline
Number: Plural nouns, verbs and pronouns are changed into single ones or the other way around. \newline
Voice: If the verbs are in active voice, they're changed to passive or the other way around. \newline
Adverb: The paraphrase has one adverb or more than the original sentence. \newline
Gender: The paraphrase differs from the original sentence in the gender of the names and pronouns. \newline
Synonym: Some words or phrases of the original sentence are replaced with synonym words or phrases. Changes in the names of people are also considered a synonym change. Classify your answers into Tense, Number, Voice, Adverb, Gender, and Synonym. \newline

Input: You definitely want to be very careful, especially if you have charitable beneficiaries. The viewing begins at 10 a.m. followed by the service at noon. \newline
Output: Adverb

\end{tcolorbox}
\end{scriptsize}
\caption{Fixed prompt (Demonstration set) for evaluation of Input-corrupted \ICIL, Example 1}
\label{fig:i3c_seed1}
\end{figure*}


\begin{figure*}
\begin{scriptsize}
\begin{tcolorbox}
Definition: In this task, you are given the name of an Indian food dish. You need to classify the dish as "sweet" or "spicy".  \newline

Input: The new UFC...  \newline
Output: sweet \newline

Definition: In this task, you are given two natural language statements with similar wording. You must choose the statement that makes less sense based on common sense knowledge. A ' ' separates the statements. Use "first" or "second" to indicate which sentence makes less sense.  \newline

Input: (Photo by Mark Rightmire, Orange County Register/SCNG) That's what a fighter's mindset is. \newline 
Output: first \newline

Definition: In this task, you are given a public comment from online platforms. You are expected to classify the comment into two classes: sexual-explicit and non-sexual-explicit. A comment is considered sexual-explicit if it explicitly portrays sexual matters. \newline

Input: u201cWere pushing back, the official said, speaking on condition of anonymity. Ripple Effect\textbackslash nJuly 1, 6 p.m. Our state has incredible people thats about moving forward. \newline
Output: Non-sexual-explicit \newline

Definition: In this task, you are given a sentence. You must judge whether a single noun or verb has been replaced with another word with the same part of speech. The inversion would result in the sentence sounding unnatural, So unnatural sentences will be considered changed. Label the instances as "Original" or "Changed" based on your judgment. \newline 

Input: In the top of the third, Jesus Pulido led off with a single before Roberto Cabrera moved him over to second base off a sacrifice bunt. \newline
Output: Original \newline

Definition: You are given a short poem which is written by a kid. Your task is to predict if the kid is in elementary school or high school. Anyone who is from grade 1st to 6th-grade will be considered as in elementary school, and 7th to 12th-grade kids will be considered as in high school. There are only two possible outputs, i.e., elementary and high. All inputs have at least one output and it cannot have both categories at a time. \newline

Input: The World Herald reports Greater Omaha Packing is set this morning to load a test run of beef for shipment to a customer in China. \newline
Output: elementary \newline

Definition: In this task, you are given a text from a social media post. Your task is to classify the given post into two categories: 1) yes if the given post is intentionally offensive, 2) no, otherwise. Also, generate label 'no' for offensive statements that appear to be unintentional, or non-offensive statements that are underhandedly offensive. Warning: the examples and instances may contain offensive language. \newline

Input: State reduces company's fine for failing to plug 2 wells\textbackslash nKENAI, Alaska (AP) He found a way to connect with everybody here, Francona said. \newline
Output: Yes \newline

Definition: In this task, you are given a public comment from online platforms. You are expected to classify the comment into two classes: toxic and non-toxic. Toxicity is defiend as anything that is rude, disrespectful, or unreasonable that would make someone want to leave a converation. \newline

Input: Click to enlarge .THURSDAY By Counseling LOL. ADVERTISEMENT Mazda6 Edwards- Nice! \#TruthRISING. -All lnorden@mww.com 2. France:\textbackslash n15. 3. ... Image jlinville@mtairynews.com NIGHT... .THURSDAY Against: L. Email: OK iBooks Twitter\textbackslash nWhatsApp\textbackslash nGoogle\textbackslash nTumblr\textbackslash nLinkedIn\textbackslash nSkype Sunny. Getty Related ... \_\_\_ lbeamish@postmedia.com WhatsApp .MONDAY Industrial ... ... .THURSDAY [1.9.6\_1107] \* ... Sunny. 1. 8. NIGHT... info@newyorkclassical.org 9/15 485 \newline
Output: Non-toxic \newline

Definition: Given a comment text in Malayalam, classify the comment into one of these categories (i) Hope speech, (ii) Not Hope Speech or (iii) Not in Expected Language. From the world of acting, Oscar-winner Mark Rylance and Moonlight star Naomie Harris were among those honored this year. \newline  

Input: u201cIve tried to be a good person,  a good parent and husband. \newline
Output: Hope Speech
\end{tcolorbox}
\end{scriptsize}
\caption{Fixed prompt (Demonstration set) for evaluation of Input-corrupted \ICIL, Example 2}
\label{fig:i3c_seed2}
\end{figure*}


\begin{figure*}
\begin{scriptsize}
\begin{tcolorbox}
Definition: You will be given a topic and an argument. Decide the argument's stance towards that topic. The argument's stance is in favor or against the topic. If the argument supports that topic, answer with "in favor"; otherwise, if the argument opposes the topic, answer with "against". \newline

Input: Her punishment didnt end there. School will resume on Monday. \newline
Output: in favor \newline

Definition: You are given an array of integers, check if it is monotonic or not. If the array is monotonic, then return 1, else return 2. An array is monotonic if it is either monotonically increasing or monotonocally decreasing. An array is monotonically increasing/decreasing if its elements increase/decrease as we move from left to right  \newline

Input: Cal Poly botany professors Matt Ritter, left, and Jenn Yost discuss the characteristics of a plant with their students. \newline
Output: 1 \newline

Definition: Given a sentence, judge the quality of this sentence by indicating "Good" and "Bad". The quality depends on the grammar and the meaning of the sentence. If a sentence is easily understandable, and doesn't have grammatical errors, answer with "Good", otherwise answer with "Bad". \newline

Input: READ: But they suffered their first defeat in all competitions since January when they were beaten, in controversial circumstances, by Middlesbrough last weekend. \newline
Output: Good \newline

Definition: We would like you to assess the QUALITY of each of the following argument (discussing Gun Control) and determine if the argument is Valid or Invalid. A valid argument is clearly interpretable and either expresses an argument, or a premise or a conclusion that can be used in an argument for the topic of gun control. An invalid argument is a phrase that cannot be interpreted as an argument or not on the topic of gun control. \newline

Input: You'll be waking up a little more awesome tomorrow. Sitting opposite The Roman Forum, the hotel is within walking distance of many key sites, including the Colosseum and the Domus Aurea. \newline 
Output: Valid \newline

Definition: Classify the given tweet into the three categories: (1) 'Hate Speech', (2) 'Offensive' and (3) ' Neither'. ' Hate Speech' is kind of a threating statement or sometimes include call for violence while 'offensive' statement just offensds someone. ' Neither' is when it doesn't fall into Hate Speech or Offensive category. \newline

Input: So, how is the city able to pay for things? \newline
Output: Hate Speech \newline

Definition: In this task, you are given a news article. Your task is to classify the article to one out of the four topics 'World', 'Sports', 'Business', 'Sci/Tech' if the article's main topic is relevant to the world, sports, business, and science/technology, correspondingly. If you are not sure about the topic, choose the closest option. Note that URLs in the text have been replaced with [Link]. \newline

Input: By essentially eliminating primaries, California reformers reasoned, they would eliminate the extremism that primaries produce. Steps taken so far by Qatar -- such as signing a memorandum of understanding on counterterrorism with the U.S. and cutting funding to Hamas -- are good steps but they need to do more, he added. Their relationship is, ultimately, symbiotic: neither can or will a company thrive in a failing society, nor can a society prosper without a successful expanding economy. \newline
Output: Sci/Tech \newline

Definition: In this task, you are given Twitter posts. Your task is to label the post's emotion (as expressed by the user) as sadness, joy, love, anger, fear, or surprise. \newline

Input: And the other hand is that because a lot of companies become listed companies, so we see market opportunity, because more and more companies begin to pay attention to their brand building. \newline
Output: fear \newline

Definition: You are given a question. You need to detect which category better describes the question. A question belongs to the description category if it asks about description and abstract concepts. Entity questions are about entities such as animals, colors, sports, etc. Abbreviation questions ask about abbreviations and expressions abbreviated. Questions regarding human beings, description of a person, and a group or organization of persons are categorized as Human. Quantity questions are asking about numeric values and Location questions ask about locations, cities, and countries. Answer with "Description", "Entity", "Abbreviation", "Person", "Quantity", and "Location". \newline

Input: He found a way to connect with everybody here, Francona said. \newline
Output: Person 
\end{tcolorbox}
\end{scriptsize}
\caption{Fixed prompt (Demonstration set) for evaluation of Input-corrupted \ICIL, Example 3}
\label{fig:i3c_seed3}
\end{figure*}

\begin{table*}
\centering
\caption{Different types of demonstrations with perturbed instruction, input, or output.}
\adjustbox{max width=\textwidth}{
\renewcommand{\arraystretch}{1.2}
\begin{tabular}{ll}
\hline
\begin{tabular}[c]{@{}l@{}}\textit{Demos}\\ \textit{of \ICIL}\end{tabular}          & \begin{tabular}[c]{@{}l@{}}(\textit{Instruction \cmark \hspace{0.1cm} Input \cmark \hspace{0.1cm} Output \cmark)}\\ Definition: In this task, you are given a dialogue from a conversation\\ between an agent and a customer. Your task is to determine the \\ speaker of the dialogue. Answer with "agent" or "customer".\\ \\ Input: I have successfully booked your ticket with flight-1017, \\ have a safe journey.\\ Output: agent\end{tabular}  

\\
\hline
\begin{tabular}[c]{@{}l@{}}\textit{Demos}\\ \textit{of Random Inst.}\end{tabular}  & \begin{tabular}[c]{@{}l@{}}(\textit{Instruction \xmark \hspace{0.1cm} Input \cmark \hspace{0.1cm} Output \cmark)}\\ \textcolor{red}{Definition: Floyd Mayweather's bout with Conor McGregor will be} \\ \textcolor{red}{"the biggest fight ever", according to UFC president Dana White.} \\ \textcolor{red}{Mariota saw his first two seasons end prematurely with injuries.} \\ \textcolor{red}{Its about taking calculated risks.}\\ \\ Input: I have successfully booked your ticket with flight-1017, \\ have a safe journey.\\ Output: agent\end{tabular} \\
\hline
\begin{tabular}[c]{@{}l@{}}\textit{Demos}\\ \textit{of Random Input}\end{tabular}  & \begin{tabular}[c]{@{}l@{}}(\textit{Instruction \cmark \hspace{0.1cm} Input \xmark \hspace{0.1cm} Output \cmark)}\\ Definition: In this task, you are given a dialogue from a conversation\\ between an agent and a customer. Your task is to determine the \\ speaker of the dialogue. Answer with "agent" or "customer".\\ \\ \textcolor{red}{Input: They were taken the Beaufort County Detention Center}\\ \textcolor{red}{and given a \$100,000 bond.}\\ Output: agent\end{tabular}                                      \\
\hline
\begin{tabular}[c]{@{}l@{}}\textit{Demos}\\ \textit{of Random Output}\end{tabular} & \begin{tabular}[c]{@{}l@{}}(\textit{Instruction \cmark \hspace{0.1cm} Input \cmark \hspace{0.1cm} Output \xmark)}\\ Definition: In this task, you are given a dialogue from a conversation\\ between an agent and a customer. Your task is to determine the \\ speaker of the dialogue. Answer with "agent" or "customer".\\ \\ Input: I have successfully booked your ticket with flight-1017, \\ have a safe journey.\\ \textcolor{red}{Output: osteology}\end{tabular} \\
\hline
\end{tabular}
}
\label{table:random_example}
\end{table*}


\end{document}